\documentclass[printer]{tADR2e}

\usepackage{graphicx}

\usepackage{amsmath}

\usepackage[ruled,vlined]{algorithm2e}

\usepackage{multirow}

\usepackage{subfigure}

\usepackage{comment}

\begin{document}

\jvol{33} \jnum{3-4} \jyear{2019} \jmonth{}
\beginpage{153}
\endpage{168}
\received{Received 29 Apr 2018, Accepted 17 Nov 2018, Published online: 05 Dec 2018}

\title{Semantic segmentation of trajectories with improved agent models
for pedestrian behavior analysis}

\author{
Toru Tamaki,
Daisuke Ogawa,
Bisser Raytchev,
Kazufumi Kaneda\\
Department of Information Engineering,
Faculty of Engineering,
Hiroshima University\\
Higashi-Hiroshima, Hiroshima 739-8527 Japan
}

\newcommand{\argmax}{\mathop\mathrm{argmax}\limits}
\newcommand{\argmin}{\mathop\mathrm{argmin}\limits}

\maketitle

\begin{abstract}

In this paper, we propose a method for semantic segmentation of pedestrian trajectories
based on pedestrian behavior models, or agents.
The agents model the dynamics of pedestrian movements in two-dimensional space using a linear dynamics model
and common start and goal locations of trajectories.
First, agent models are estimated from the trajectories obtained from image sequences.
Our method is built on top of the Mixture model of Dynamic pedestrian Agents (MDA); however, the MDA's trajectory modeling and estimation are improved.
Then, the trajectories are divided into semantically meaningful segments.
The subsegments of a trajectory are modeled by applying a hidden Markov model using the estimated agent models.
Experimental results with a real trajectory dataset
show the effectiveness of the proposed method
as compared to the well-known classical Ramer-Douglas-Peucker algorithm
and also to the original MDA model.
\end{abstract}

\section{Introduction}

The analysis of the behavior and trajectories of pedestrians captured by video cameras is
an important topic in the computer vision field, and has been widely studied over the decades \cite{yi2015,Yi2016TIP,Yi2016ECCV,Alahi2016CVPR,Lee2017CVPRdesire,Morris2008TCSVT}.
When researchers handle trajectories,
they frequently perform segmentation to reduce the computation cost
and extract local information. There are three typical approaches \cite{Zheng,Feng2016survey}:
\begin{itemize}
 \item Temporal segmentation: a trajectory is split at the points at which two observed locations are temporally at a distance from each other.
 \item Shape-based segmentation: a trajectory is split at points of larger curvature, which indicate that the target may change its direction at that point. This is used for simplifying the shape of trajectories; the Ramer-Douglas-Peucker (RDP) algorithm \cite{Ramer,Douglas-Peucker} is a well-known approach of this type.
 \item Semantic segmentation: a whole trajectory is divided into semantically meaningful segments; many methods have been proposed for different tasks \cite{Yuan,Lee,Zheng2008a,Zheng2008b,Zheng2010}.
\end{itemize}
In this paper, we focus on the third approach, semantic segmentation of trajectories based on models of human behavior (or agents).
We represent the semantics of segments using agent models to capture the direction in which pedestrians walk when entering and exiting a scene. 
Hence, our task is to segment a trajectory into several different sub-trajectories represented using different agent models.
It would be beneficial if segments associated behavior \cite{Morris2011TPAMI,Li2015survey} could be obtained;
for example, a long-term temporal change of behavior
from a sequence of segments may be found, which is not possible using a series of raw trajectory coordinates. Furthermore, collective analysis by using many trajectory segments with associated behavior may be useful to find potential crowded regions in a scene.
However, no segmentation methods of trajectories have been proposed for facilitating the task of human behavior analysis.
Our proposed method first estimates agent models by using the Mixture model of Dynamic pedestrian Agents (MDA) \cite{MDA}, and then segments trajectories using the learned agent models by applying hidden Markov models (HMM) \cite{Baum,Viterbi}.
\footnote{A conference version of this paper was presented in \cite{Ogawa2018FCV}. This paper extends that version with the extension of the method description, and more extensive evaluations.}

\section{Related work}

The RDP algorithm \cite{Ramer,Douglas-Peucker} is frequently used for
trajectory simplification. It segments a trajectory while preserving important points in order to
retain the trajectory's shape to the greatest extent possible. First, the start and end points of a trajectory are preserved.
Then, RDP finds the point located the greatest distance away from the line between two preserved points
and keeps it if the distance is larger than threshold $\epsilon$.
This process iterates recursively until no further points are preserved.
Finally, all the preserved points are used to segment the trajectory.
This method is simple and preserves the approximate shape of the trajectory;
however, an appropriate value of $\epsilon$ has to be specified.

Task oriented methods have also been proposed.
Yuan et al. \cite{Yuan} proposed a system called T-Finder, which recommends
to taxi drivers locations at which as many potential customers as possible exist,
and to users locations where they can expect to find taxis.
For this purpose, these methods estimate the locations of taxis based on their driving trajectories
and segment the trajectories as a pre-processing procedure.
Lee et al. \cite{Lee} proposed the trajectory outlier detection algorithm (TRAOD), an algorithm for finding outliers in trajectories
based on segmentation by using the minimum description length (MDL) principle.
Zheng et al. estimated transportation modes \cite{Zheng2008a,Zheng2008b,Zheng2010},
such as walking, car, bus, and bicycle, and used them for semantic segmentation in terms of the mode of transportation.

In contrast, our proposed method uses semantic human behavior models, called agent models,
learned from pedestrian trajectories in videos.
This task is entirely different from that of the RDP algorithm, in which merely the simplified shape of
a trajectory is taken into account and segments have no relation to behavior models.
Furthermore the objective of our study differed from that of previous studies on semantic segmentation of trajectories \cite{Zheng2008a,Zheng2008b,Zheng2010}, in which a trajectory is composed of
different vehicles. Our goal was to perform the segmentation of a person's trajectory
by dividing it into different behavior (agent) models.

\section{Learning agent models}
\label{subsec:MDA}

In this section, we describe
 the MDA model \cite{MDA} proposed by Zhou et al.
and our improved model, called improved MDA (iMDA).
The MDA model is a hierarchical Bayesian model for representing
pedestrian trajectories by using a mixture model of linear dynamic systems and common start and goal locations (called beliefs).
The parameters of the dynamics and beliefs of each agent are estimated by using an expectation maximization (EM) algorithm
that alternates the E step (expectation over hidden variables) and M step (parameter estimation).
In other words, soft clustering of trajectories to the estimated agents
and optimization by using weighted sums are iterated.

It is reasonable to use MDA for the task of semantic segmentation of pedestrian trajectories,
because it estimates agent models that reflect pedestrian behaviors: the direction in which the person is walking, his/her speed, and the locations from which and to which the person is moving. These are modeled by agents using dynamics and beliefs.
However, MDA was proposed for performing clustering of trajectories, and therefore, we extend it by adding HMM for segmentation. Furthermore, the original MDA suffers a convergence problem, and hence, we propose an improved version of MDA.

\def\y{\boldsymbol{y}}
\def\x{\boldsymbol{x}}
\def\b{\boldsymbol{b}}
\let\oldmu\mu
\def\mu{\boldsymbol{\oldmu}}

\subsection{Formulation}

Let $\y_t \in R^2$ be two-dimensional coordinates at time $t$ of a pedestrian trajectory
$\y = \{ \y_0, \y_1, \ldots, \y_\tau\}$,
and $\x_t \in R^2$ be the corresponding state
of the linear dynamic system
\begin{align}
\x_t &\sim P(\x_t | \x_{t-1}) = N(\x_t | A \x_{t-1} + \b, Q) \label{eq:xt}
\\
\y_t &\sim P(\y_t | \x_t) = N(\y_t | \x_t, R) \label{eq:yt},
\end{align}
where
$N(\cdot)$ represents a normal distribution with covariance matrices $Q, R \in R^{2\times 2}$,
and $A \in R^{2\times 2}$ is the state transition matrix and $\b \in R^2$ is the translation vector.
This means that the state transition is assumed to be a similar transformation.
In this study, we explicitly used the translation vector for similar transformations,
while Zhou et la. \cite{MDA} used homogeneous coordinates for their formulation.

MDA represents pedestrian trajectories modeled by agents with 
dynamics $D$ and belief $B$.
Here, dynamics $D = (A, \b, Q, R)$ describes a pedestrian's movement in the two-dimensional scene.
Belief $B$ describes the starting point $\x_s$ and end point $\x_e$ of the trajectory,
each represented by normal distributions:
\begin{align}
\x_s &\sim p(\x_s) = N(\x_s | \mu_s, \Phi_s)
\\
\x_e &\sim p(\x_e) = N(\x_e | \mu_e, \Phi_e);
\end{align}
that is, belief is represented as $B = (\mu_s, \Phi_s, \mu_e, \Phi_e)$,
describing the common starting and end locations.
The mixture weights are written as $\pi_m = p(z = m)$
with hidden variable $z$, which indicates that the trajectory is generated by the $m$-th agent.

Trajectory observation $\y = \{ \y_0, \y_1, \ldots, \y_\tau\}$
may not start and end at the exact start and end points
$\x_s$ and $\x_e$. For example, a pedestrian is visible to the camera
and tracked over video frames to generate $\y_0, \y_1, \ldots$; however,
the common starting point $\mu_s$ may be occluded by walls or signboards.
Therefore, we model states $\x$
before and after the observed points $\y_0, \ldots, \y_\tau$ of the trajectory:
\begin{align}
\x = \{
\x_s = \x_{-t_s}, \x_{-t_s + 1}, \ldots, \x_0,
\x_1, \x_2, \ldots, \x_\tau,
\x_{\tau + 1}, \ldots, \x_{\tau + t_e} = \x_e
\}.
\end{align}
The length of the observation is $\tau+1$,
and the length of the states before the observation is $t_s$ and after the observation $t_e$.

\begin{figure}[tb]
\begin{center}
\includegraphics[width=0.8\linewidth]{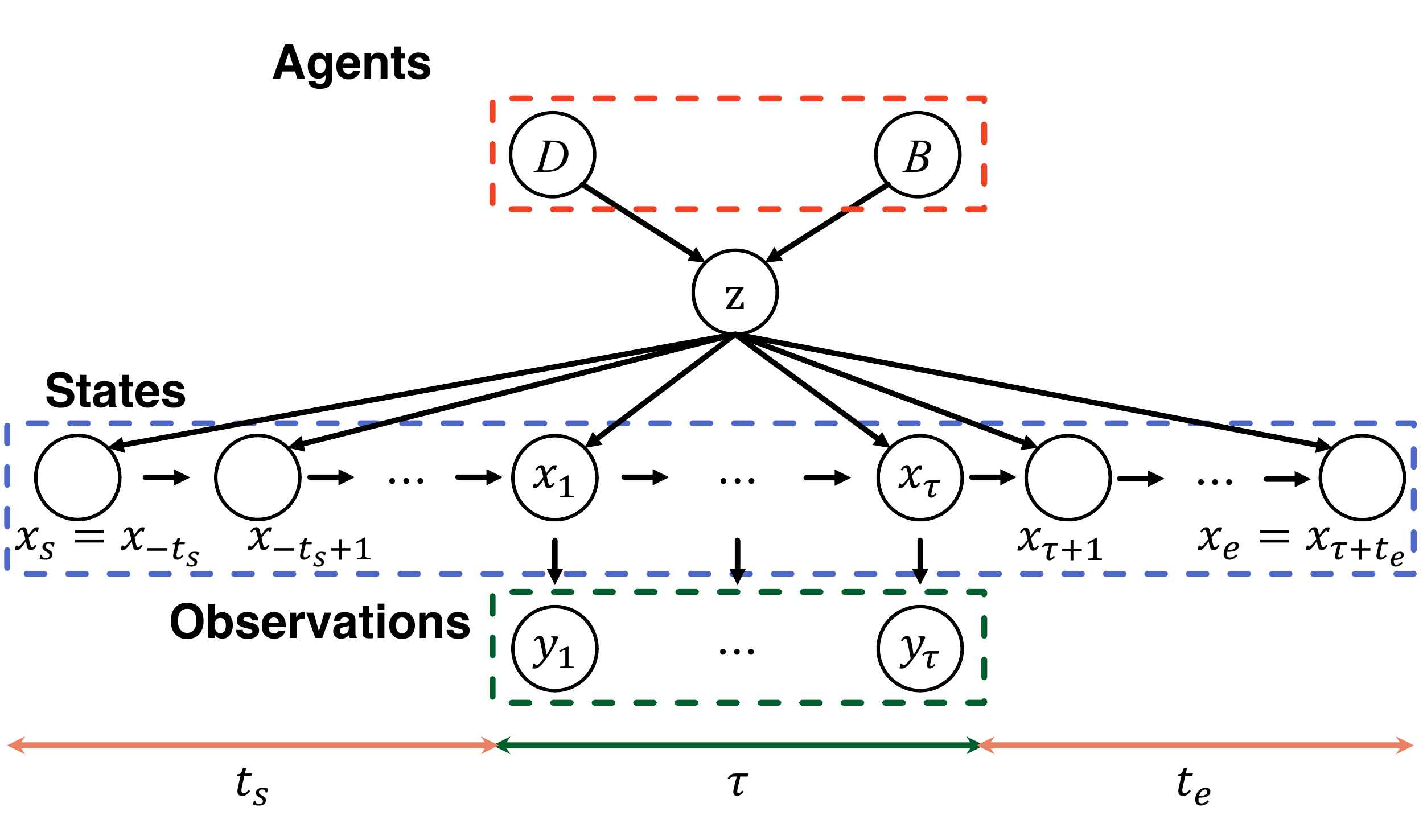}
\end{center}
\caption{Graphical model of the original MDA. \cite{MDA}}
\label{fig:1}
\end{figure}

Figure \ref{fig:1} shows a graphical model of MDA.
Observed trajectory $\y$ is generated by states $\x$,
and all states are governed by a single hidden variable $z$
that switches agent models $D$ and $B$.

\subsection{Learning}

Given $K$ trajectories $Y = \{ \y^k \}$,
where $\y^k$ is $k$-th observation,
MDA \cite{MDA} estimates
$M$ agents $\Theta = \{ (D_m, B_m, \pi_m)\}_{m=1}^M$
by maximizing the log likelihood function
\begin{align}
L = \sum_k \log p(\y^k, \x^k, z^k, t_s^k, t_e^k | \Theta),
\end{align}
where the joint probability is given by
\begin{align}
p(\y^k, \x^k, z^k, t_s^k, t_e^k)
=
p(z^k) p(t_s^k) p(t_e^k)
p(\x_s^k)
p(\x_e^k)
\prod_{t=-t_s^k+1}^{\tau^k + t_e^k} p(\x_t^k | \x_{t-1}^k)
\prod_{t=0}^{\tau^k} p(\y_t^k | \x_t^k)
\end{align}
with respect to parameters $\Theta$.

The abovementioned equations used in \cite{MDA} have many hidden variables,
and we simplify the likelihood by writing
the hidden variables as
$H = \{ h^k \}, h^k = \{ z^k, t_s^k, t_e^k \}$
as follows.
\begin{align}
L = \sum_k \log p(\y^k, \x^k, h^k | \Theta).
\end{align}
The EM algorithm is used for estimation by alternating the E and M steps,
as $H$ is not observed.

\subsubsection{E step of the original MDA}

The E step of MDA \cite{MDA} takes the expectation
of the log likelihood with respect to the hidden variables $H$:
\begin{align}
Q(\Theta, \hat\Theta)
&= E_{X, H | Y, \hat\Theta} [L]
\\
&= E_{H | Y, \hat\Theta} [ E_{X | Y, H, \hat\Theta} [L] ]
\\
&= \sum_k \sum_{h^k}
\gamma^k E_{\x^k | \y^k, h^k} [\log p(\y^k, \x^k, h^k | \Theta)]
\\
&= \sum_k \sum_{h^k}
\gamma^k \log p(\y^k, \hat\x^k, h^k | \Theta),
\end{align}
where $\hat\x^k = E_{\x^k | \y^k, h^k} [\x^k]$
is computed by using the modified Kalman filter \cite{Palma2007,MDA}.
Note that
$\hat\x^k$ should have the subscript $h^k$, because
it differs for different agents $z=m$; however, we omit it for simplicity.

Weights $\gamma^k$ are posterior probabilities given as
\begin{align}
\gamma^k
=
p( h^k | \y^k, \hat\Theta)
=
\frac{p( h^k | \hat\Theta)
      p( \y^k | h^k, \hat\Theta)}
      {p( \y^k | \hat\Theta)}.
\end{align}

By assuming independence among hidden variables $z$, $t_s$, and $t_e$,
we have
\begin{align}
p(h^k | \hat\Theta)
= p(z^k, t_s^k, t_e^k | \hat\Theta)
= p(z^k | \hat\Theta) p(t_s^k | \hat\Theta) p(t_e^k | \hat\Theta).
\end{align}
By removing $t_s$ and $t_e$ by assuming them to be uniform,
we have
\begin{align}
\gamma^k
&= \frac{p(z^k | \hat\Theta) p( \y^k | h^k, \hat\Theta)}
{\sum_{h^k} p(z^k | \hat\Theta) p( \y^k | h^k, \hat\Theta)},
\end{align}
where likelihood $p( \y^k | h^k, \hat\Theta)$ is also computed
by using the modified Kalman filter \cite{Palma2007,MDA}.

\subsubsection{E step of improved MDA}

The E step of MDA described above
suffers a convergence problem in practice,
because it does not  explicitly take belief $B$ into account in $\gamma$.
In fact, it is implicitly included in the form of $p( \y^k | h^k, \hat\Theta)$; however, the modified (or ordinal) Kalman filter does not deal with belief parameters.

We solve the problem by introducing two improvements.
First, we separate $\x_s$ and $\x_e$ from the other states and
use them as hidden variables. This is because these starting and end states
are in fact hidden states when $t_s$ and $t_e$ are non-zero (as is usual).
Belief parameters are affected by these starting and end states only,
and therefore, it is necessary to include them as the information of beliefs $B$ in the E step.
Hereafter, $\x_{1:T}$ denotes the sequence of states $\x$, except $\x_s$ and $\x_e$.

\begin{figure}[t]
\centering
\includegraphics[width=.48\linewidth]{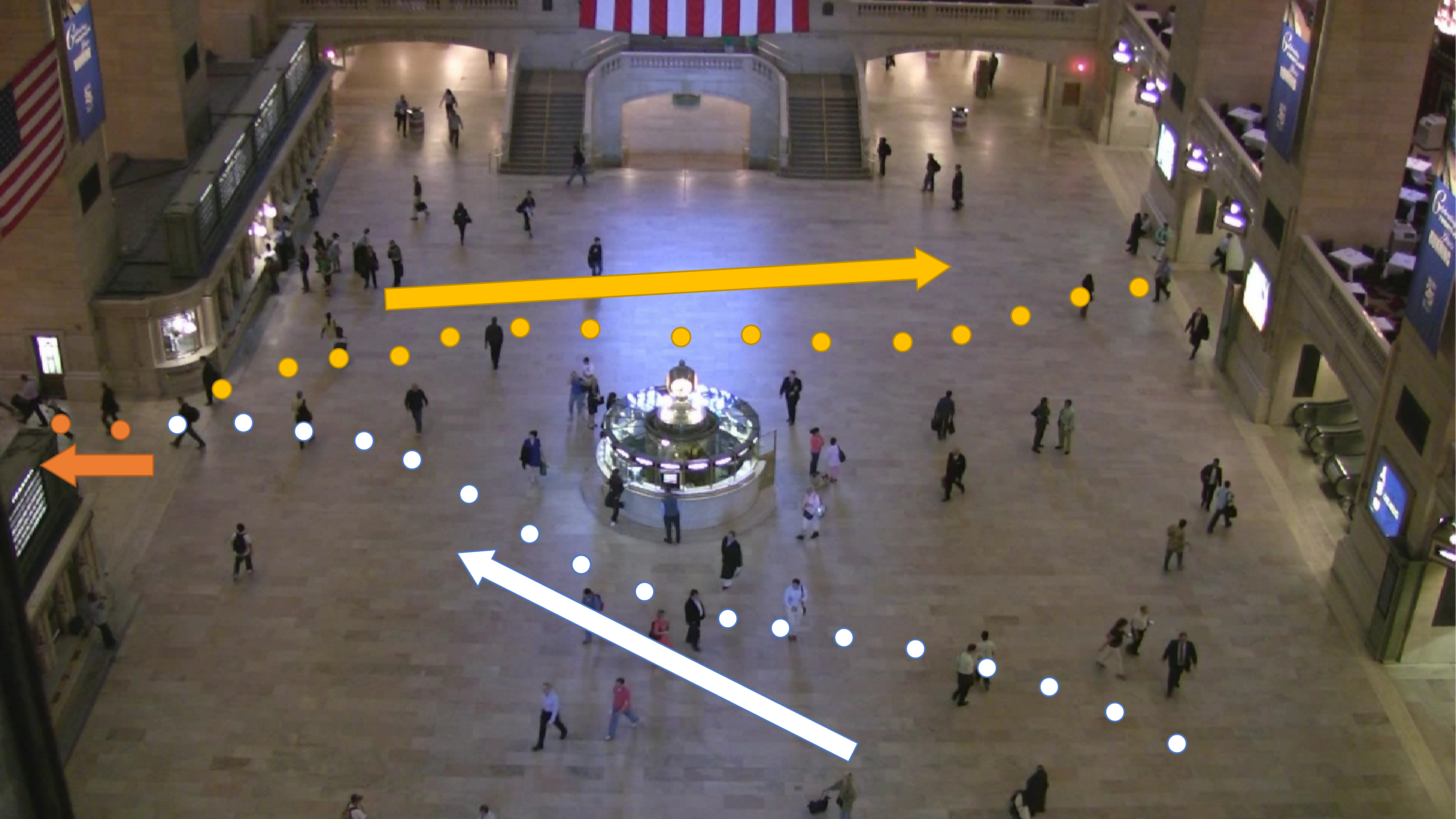}
\caption{

States after the observations for the case of the left exit are shown in orange, and the states for the case of the right exit are shown in
yellow. The image is from \cite{yi2015}.}

\label{fig:why_poisson}
\end{figure}

Second, we explicitly model $t_s$ and $t_e$ with Poisson distribution:
\begin{align}
t_s &\sim p(t_s = k) = \frac{\lambda_s^k}{k!} e^{-\lambda_s}
\\
t_e &\sim p(t_e = k) = \frac{\lambda_e^k}{k!} e^{-\lambda_e}.
\end{align}
Uniform distributions are assumed in the E step of the original MDA;
however, this may prevent the iteration from converging,
because any number of states before and after observations is allowed
with equal possibilities. This means that a trajectory can start from
a location very distant from the beginning of the observation,
although usually this does not occur.
This is illustrated in Figure \ref{fig:why_poisson}.
An example trajectory observation is shown as dots in white and starts from the right bottom of the scene and ends at the left.
This trajectory is expected to go out from the left exit because there are no further observations.
The uniform distribution assigns the same probabilities to
both the cases of the left and right exits;
however, the latter case is much less likely to happen. In contrast,
the Poisson distribution reasonably assigns a higher probability to
the case of the left exit.

As in the E step above, we take the expectation with hidden variables,
including $X_s = \{\x_s^k\}$ and $X_e = \{\x_e^k\}$:
\begin{align}
Q(\Theta, \hat\Theta)
&= E_{X_{1:T}, X_s, X_e, H | Y, \hat\Theta} [L]
\\
&= E_{H, X_s, X_e | Y, \hat\Theta} [ E_{X_{1:T} | Y, H, X_s, X_e, \hat\Theta} [L] ]
\\
&\simeq \sum_k \sum_{h^k}
\gamma^k E_{\x_{1:T}^k | \y^k, \x_s^k, \x_e^k, h^k} [\log p(\y^k, \x_{1:T}^k, \x_s^k, \x_e^k, h^k | \Theta)]
\\
&= \sum_k \sum_{h^k}
\gamma^k \log p(\y^k, \hat\x_{1:T}^k, \x_s^k, \x_e^k, h^k | \Theta).
\end{align}
There is an approximation in the third line,
because marginalizing with respect to $\x_s$ and $\x_e$ is computationally expensive.
The effect is, however, negligible, because usually
the differences between next states are very small:
the difference between $\x_s = \x_{-t_s}$ and $\x_{-t_s + 1}$,
and the difference between $\x_e = \x_{\tau + t_e}$ and $\x_{\tau + t_e - 1}$
are smaller than the distances to the locations of start $\mu_s$ and goal $\mu_e$.

We further approximate it by omitting $\x_s$ and $\x_e$:
\begin{align}
Q(\Theta, \hat\Theta)
&= \sum_k \sum_{h^k}
\gamma^k \log p(\y^k, \hat\x_{1:T}^k, h^k | \Theta).
\end{align}
The effect of this omission on $Q$ is very small because only the first and last states from a long sequence of states exist.
We compute $\hat\x^k = E_{\x_{1:T}^k | \y^k, h^k} [\x^k]$
(as an approximation, again by omitting $\x_s$ and $\x_e$)
by using the modified Kalman filter \cite{Palma2007,MDA}, as in the original E step.

There are two differences between $Q$ above and that in the original E step.
First, the original $Q$ has all states $\x$,
but our improved version of $Q$, shown above, has $\x_{1:T}$ (except $\x_s$ and $\x_e$).
Second, weights $\gamma$ are different.
Here, we derive $\gamma^k$ as
\begin{align}
\gamma^k
&=
p( h^k, \x_s^k, \x_e^k | \hat\x_{1:T}^k, \y^k, \hat\Theta)
\\
&=
\frac{p( h^k, \x_s^k, \x_e^k | \hat\x_{1:T}^k, \hat\Theta)
      p( \y^k | h^k, \x_s^k, \x_e^k, \hat\x_{1:T}^k, \hat\Theta)}
      {p( \y^k | \hat\x_{1:T}^k, \hat\Theta)}
\\
&=
\frac{p( h^k | \hat\Theta)
      p( \x_s^k | \hat\Theta)
      p( \x_e^k | \hat\Theta)
      p( \y^k| h^k, \x_s^k, \x_e^k, \hat\x_{1:T}^k, \hat\Theta)}
      {p( \y^k | \hat\x_{1:T}^k, \hat\Theta)}
\\
&=
\frac{p(z^k | \hat\Theta) p(t_s^k | \hat\Theta) p(t_e^k | \hat\Theta)
      p( \x_s^k | \hat\Theta)
      p( \x_e^k | \hat\Theta)
      p( \y^k| h^k, \x_s^k, \x_e^k, \hat\x_{1:T}^k, \hat\Theta)}
      {\sum_{h^k}
      p(z^k | \hat\Theta) p(t_s^k | \hat\Theta) p(t_e^k | \hat\Theta)
      p( \x_s^k | \hat\Theta)
      p( \x_e^k | \hat\Theta)
      p( \y^k| h^k, \x_s^k, \x_e^k, \hat\x_{1:T}^k, \hat\Theta)
      }
\\
&\simeq
\frac{p(z^k | \hat\Theta) p(t_s^k | \hat\Theta) p(t_e^k | \hat\Theta)
      p( \x_s^k | \hat\Theta)
      p( \x_e^k | \hat\Theta)
      p( \y^k| h^k, \hat\x_{1:T}^k, \hat\Theta)}
      {\sum_{h^k}
      p(z^k | \hat\Theta) p(t_s^k | \hat\Theta) p(t_e^k | \hat\Theta)
      p( \x_s^k | \hat\Theta)
      p( \x_e^k | \hat\Theta)
      p( \y^k| h^k, \hat\x_{1:T}^k, \hat\Theta).
      }
\end{align}
Note that we assume independence among $\x_s^k$, $\x_e^k,$ and $h^k$; however,
 $t_s$ and $t_e$ are modeled by Poisson distribution,
and hence, $p(t_s)$ and $p(t_e)$ remain.
In the last line, we approximate it again by omitting $\x_s$ and $\x_e$
for computing the modified Kalman filter.

\subsubsection{M step of improved MDA}

In the M step, we find
$\hat\Theta = \mathrm{arg}\max_\Theta Q(\Theta, \hat\Theta)$
by solving a system of equations
obtained by differentiating $Q$ with respect to $\Theta$.

In the following formulas of the proposed iMDA, we introduce two improvements.
First, we derive the formulas for the parameters of Poisson distribution
$\lambda_s$ and $\lambda_e$.
Second, in fact the MDA formulas (particularly for $A$ and $\b$) are incorrect and we show
the correct formula with its derivation in the Appendix. In the MDA formulation,
homogeneous coordinates are used for similar transformation
with a 3$\times$3 matrix, which is not useful for differentiation.
In our formulation, we use explicitly translation vector $\b$ instead of homogeneous coordinates, which leads to
\def\vec{\mathrm{vec}}
\begin{align}
\begin{pmatrix}
\vec(\hat{A}_m^T) \\
\hat{\b_m}
\end{pmatrix}
&=
\left(\sum_{k,h/z=m,t}\gamma^k
\begin{pmatrix}
(I_2 \otimes {\hat{\x}_{t-1}^k} (\hat{\x}_{t-1}^k){}^T) &
(I_2 \otimes \hat{\x}_{t-1}^k) \\
(I_2 \otimes (\hat{\x}_{t-1}^k){}^T) & I_2
\end{pmatrix}
\right)^{-1}\notag\\
&\phantom{=}
\left(\sum_{k,h/z=m,t}\gamma^k
\begin{pmatrix}
\vec(\hat{\x}_t^k (\hat{\x}_{t-1}^k){}^T)\\
\hat\x_{t}
\end{pmatrix}
\right)
\end{align}

\begin{align}
\hat{Q}_m
&=
\frac{\sum_{k,h/z=m,t}\gamma^k
	(\hat\x_{t}^k - \hat{A}_m \hat\x_{t-1}^k - \hat{\b})
	(\hat\x_{t}^k - \hat{A}_m \hat\x_{t-1}^k - \hat{\b})^T}
	{\sum_{k,h/z=m,t} T \gamma^k}
\\
\hat{R}_m
&=
\frac{\sum_{k,h/z=m,t}\gamma^k
(\y_{t}^k - \hat\x_{t}^k)
(\y_{t}^k - \hat\x_{t}^k){}^T
}
	{\sum_{k,h/z=m,t}\gamma^k}
\\
\hat\mu_{sm}
&=
\frac{\sum_{k,h/z=m}\gamma^k \hat\x_s^k}
	{\sum_{k,h/z=m}\gamma^k}
\\
\hat\mu_{em}
&=
\frac{\sum_{k,h/z=m}\gamma^k \hat\x_e^k}
	{\sum_{k,h/z=m}\gamma^k}
\\
\hat\Phi_{sm}
&=
\frac{\sum_{k,h/z=m}\gamma^k (\hat\x_s^k - \hat\mu_{sm})(\hat\x_s^k - \hat\mu_{sm})^T}
	{\sum_{k,h/z=m}\gamma^k}
\\
\hat\Phi_{em}
&=
\frac{\sum_{k,h/z=m}\gamma^k (\hat\x_s^k - \hat\mu_{em})(\hat\x_s^k - \hat\mu_{em})^T}
	{\sum_{k,h/z=m}\gamma^k}
\\
\hat\pi_m
&=
\frac{\sum_{k,h/z=m}\gamma^k}
	{\sum_{k,h}\gamma^k}
	= p(z^k = m)
\\
\lambda_{sm}
&=
\frac{\sum_{k,h/z=m} \gamma^k t_s^k}{\sum_{k,h/z=m} \gamma^k}
\\
\lambda_{em}
&=
\frac{\sum_{k,h/z=m} \gamma^k t_e^k}{\sum_{k,h/z=m} \gamma^k}.
\end{align}
Here, $\vec$ is the vectorization operator
and $\otimes$ is the tensor product.
Notation $h/z=m$ means that $z$ is fixed to $m$ in the summation
over $h = \{z, t_s, t_e\}$.
The ranges of summation are
from 1 to $K$ for trajectory $k$,
from 1 to $M$ for agent $m$,
and from 1 to $\tau$ for time $t$, if observation $\y_t$ is involved, and
otherwise from $-t_s$ to $\tau + t_e$, i.e., between the start and end points of the observations.
Note that the search range of $t_s$ and $t_e$
is reduced by an ad hoc technique used in \cite{MDA}.

\def\srho{\rho}
\let\oldrho\rho
\def\rho{\boldsymbol{\oldrho}}
\def\A{\boldsymbol{A}}
\def\B{\boldsymbol{B}}
\def\X{\boldsymbol{X}}
\def\Y{\boldsymbol{Y}}
\def\Z{\boldsymbol{Z}}
\def\v{\boldsymbol{v}}

\begin{figure}[tb]
\begin{center}
\includegraphics[width=\linewidth]{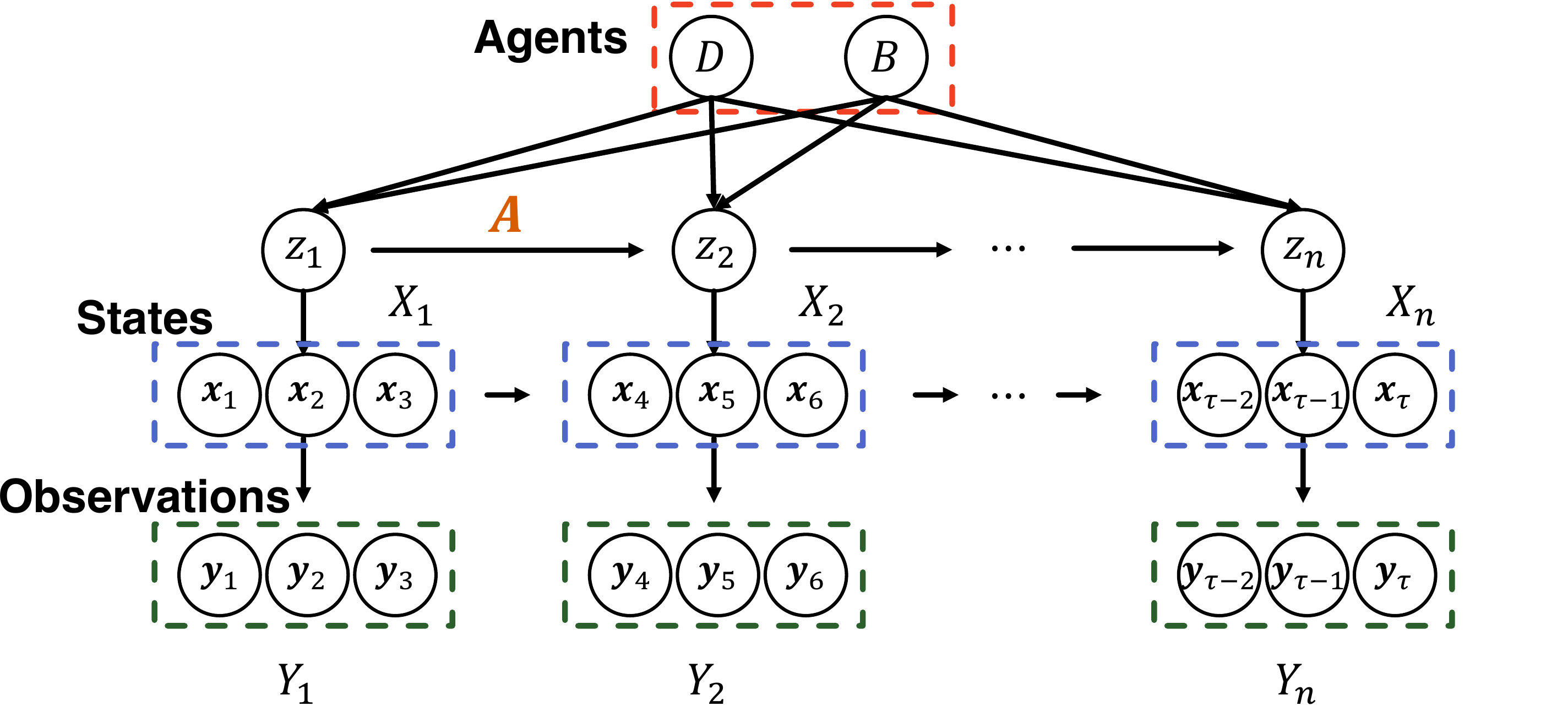}
\end{center}
\caption{Graphical model of the proposed method.}
\label{fig:5}
\end{figure}

\section{Trajectory segmentation with agent models}

Figure \ref{fig:5} shows the graphical model
of the proposed trajectory segmentation.
We propose using agent models obtained by iMDA
for segmentation with HMM.
In contrast to the MDA model (Fig. \ref{fig:1}) that shares the hidden variable $z$ across all states $\x$, our model has different hidden variables $\Z = \{z_1, z_2, \ldots\}$ indicating
for which agent model the state is generated.
However, it is difficult to use a single two-dimensional point for inferring the agent to which the point
belongs, because agent models represent dynamics and beliefs, which are difficult to infer using a single point. Instead, we use successive $N$ states in the MDA model ($\x_1, \x_2, \x_3$ in Fig. \ref{fig:5} as $N=3$ is used in the experiments)
as a single state $\X_1$ corresponding to a hidden variable $z_1$.
Note that in Fig. \ref{fig:5} we collect three successive states without overlapping; however,
the following discussion is effective without modification for the overlapping case
(e.g., $\x_1, \x_2, \x_3$ and $\x_2, \x_3, \x_4$, and so on).

Our model, shown in Fig. \ref{fig:5}, is considerably more complicated than the MDA model
and it is difficult to learn all the parameters of HMM and iMDA jointly.
Instead, we propose a two-stage algorithm composed of agent estimation followed by segmentation.
First, the agent models ($D$ and $B$) are
estimated with the iMDA described in the previous section.
We denote these agents by $\Theta = \{ (D_m, B_m, \pi_m)\}_{m=1}^{M} = \{\omega_m\}_{m=1}^{M}$.
As a byproduct, states $\x_t$ in the MDA model
are also obtained through the modified Kalman filter as
$\hat\x^k = E_{\x_{1:T}^k | \y^k, h^k} [\x^k]$ for each agent $m$.
Therefore, we use these estimated states $\hat\x_1, \hat\x_2, \ldots$ to construct
states $\X_1, \X_2, \ldots$.

Second, we fix these states $\X = \{\X_1, \X_2, \ldots\}$ during the segmentation procedure;
in other words, states $\X$ are used as observations for HMM.
For HMM training, we use the Baum-Welch algorithm \cite{Baum}
to estimate the state transition matrix $\A = \{a_{ij}\}$, an $M\times M$ matrix,
the $(i, j)$ element of which is transition probability $a_{ij} = p(z_t = j | z_{t-1} = i)$
from agent $\omega_i$ to agent $\omega_j$.
Each state $\X_t = \{ \hat\x_{t1}, \hat\x_{t2}, \ldots \hat\x_{tN}\}$
is supposed to be generated based on the output (or emission)
probability matrix $\B = \{b_{mt}\}$.
We define it as
\begin{align}
b_{mt} \sim p(\X_t | z=m) = \prod_{i=2}^{N} N(\hat\x_{ti} | \hat{A}_m \hat\x_{t,i-1} + \hat\b_m, \hat{Q}_m).
\end{align}
This is a likelihood representing how the sequence
$\hat\x_{t1}, \hat\x_{t2}, \ldots \hat\x_{tN}$ fits
the dynamics $\hat{A}_m, \hat\b_m, \hat{Q}_m$ of agent $m$.
Here, we do not use the belief parameters, because state $\X_t$ corresponds to
a short trajectory segment,
and it is not stable to find the start and goal locations from a short segment $\X_t$.
Note that we do not estimate the initial distribution of agents during the training, but
instead use weights $\{\pi_m\}$ estimated in the agent estimation.

To estimate $\A$,
the Baum-Welch algorithm performs
the EM algorithm to maximize the following log likelihood given $K$ trajectories.
\begin{align}
Q(\A, \A^{old})
= \sum_{k=1}^K E_{\Z| \X, \A^{old}} [\ln{p(\X, \Z | \A)}],
\end{align}
where
\begin{align}
p(\X, \Z | \A)
= \prod_{t=2}^n p(z_t | z_{t-1}) \prod_{t=1}^n p(\X_t | z_t).
\end{align}

When $\A$ has been estimated,
we use the Viterbi algorithm \cite{Viterbi}
for estimating hidden variables $\Z_t$ for the test trajectories.
This is a MAP estimate that maximizes the following posterior probability
of $\Z$, given new test trajectory state $\X$:
\begin{align}
\hat\Z
= \argmax_{\Z} p(\Z | \X)
= \argmax_{\Z} p(\X, \Z).
\end{align}
When a new test trajectory $\y = \{\y_0, \ldots, \y_\tau\}$ is given,
states $\hat\x = \{\hat\x_0, \ldots, \hat\x_\tau \}$
are obtained by using the modified Kalman filter
with the estimated agent models, and
then, the HMM observation sequence $\X = \{\X_t\}$ is constructed
so that $\X_1 = \{\x_0, \x_1, \x_2\}$ and so on.
After the Viterbi algorithm has been performed,
$\hat\Z = \{\Z_t\}$ is obtained and then
converted to a sequence $s = \{s_0, s_1, \ldots, s_\tau\}$ of the same length
as $\y$ as a segmentation result ($s_i \in \{1, \ldots, m\}$),
so that $s_0 = s_1 = s_2 = \Z_1$ and so on.

\section{Experiments}

We compared the proposed method, denoted by iMDA+HMM, with
the RDP algorithm \cite{Ramer,Douglas-Peucker}
in terms of segmentation accuracy.
Trajectories in the Pedestrian Walking Path Dataset \cite{yi2015}
were used for the experiments.
This dataset contains 12684 pedestrian trajectories
in videos of size $1920 \times 1080$ pixels.
We evaluated the methods using real trajectories from the dataset.

\begin{figure}[t]
\centering
\includegraphics[width=.48\linewidth]{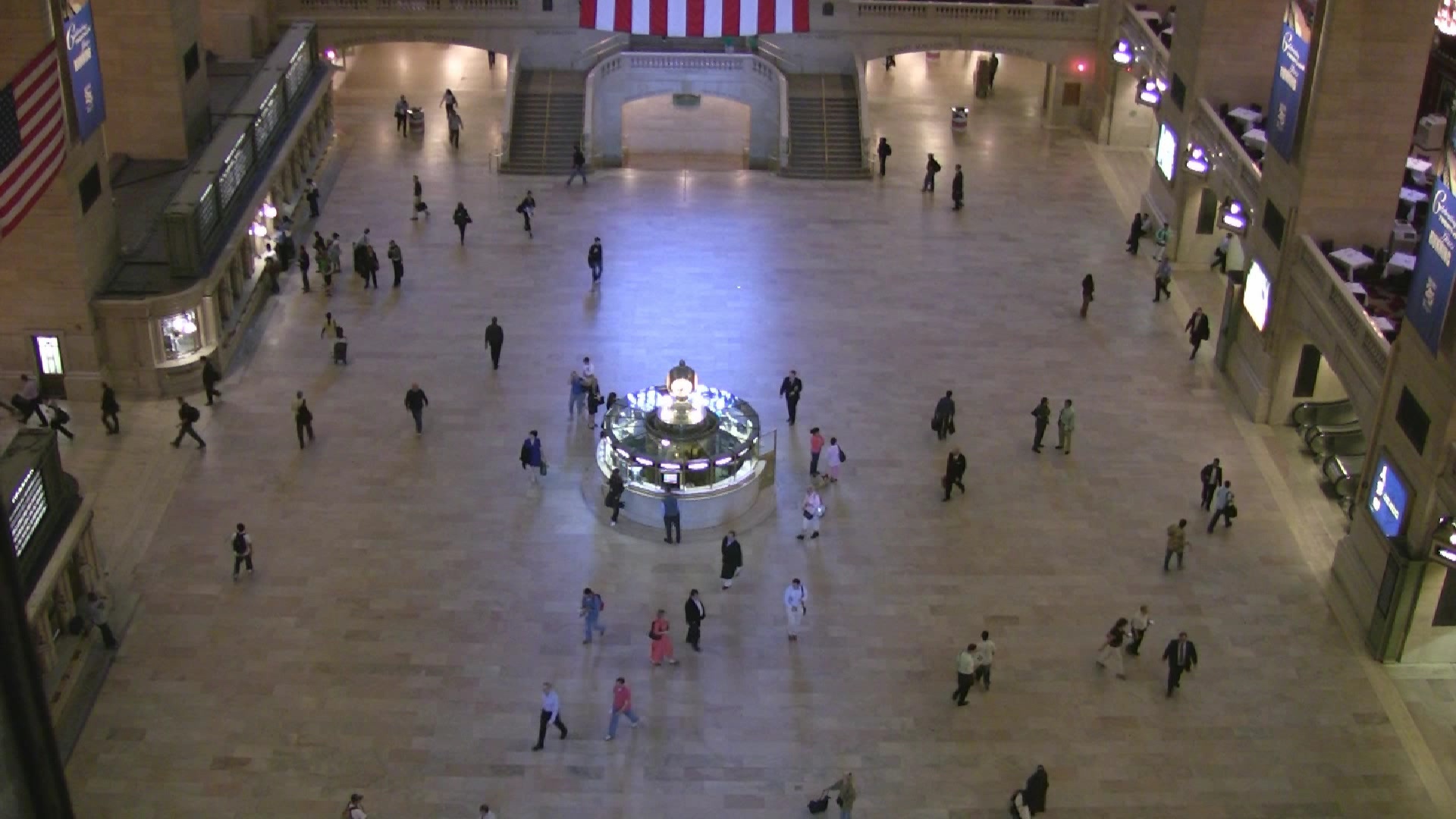}
\includegraphics[width=.48\linewidth]{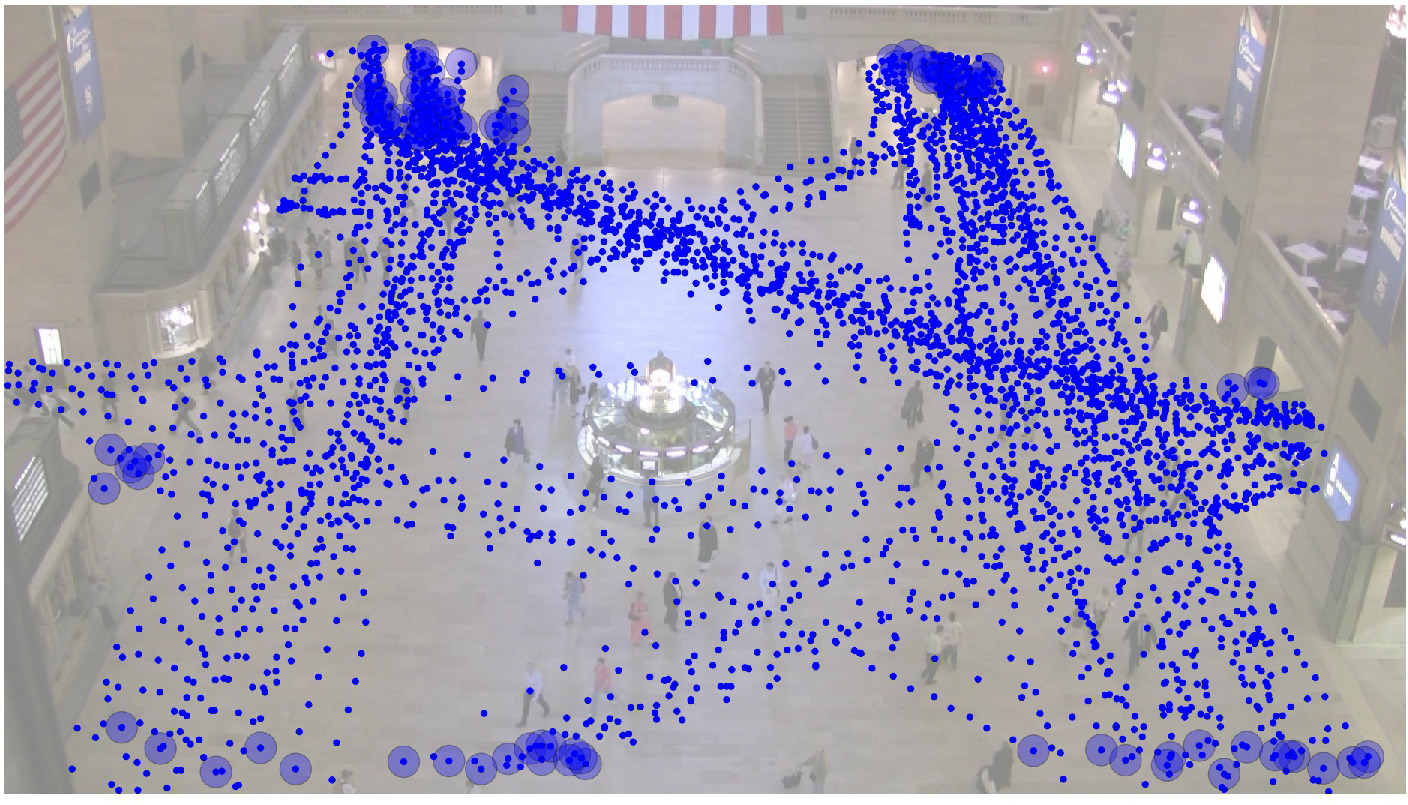}
\caption{The Pedestrian Walking Path Dataset \cite{yi2015}.
Left: typical frame. Right: one hundred randomly sampled  trajectories.
Each trajectory is shown as a series of points with a larger (faint) circle at the first point of the trajectory.}
\label{fig:dataset}
\end{figure}

\subsection{Metrics}
\label{subsec:metrics}

\begin{figure}[t]
\begin{center}
\includegraphics[width=.7\linewidth]{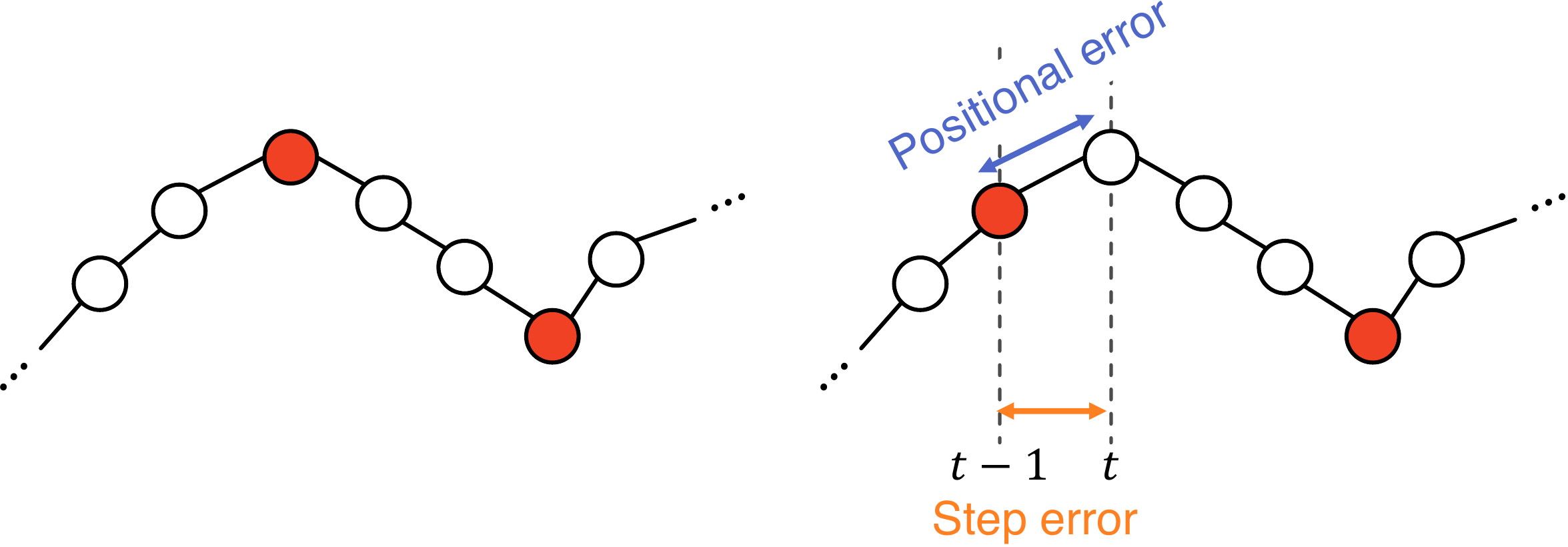}
\end{center}
\caption{Positional and step errors. Left: trajectory with ground truth segmentation points in red. Right: estimated segmentation points in red. The difference is measured by distance for positional errors and by the index of point sequence of the trajectory for step errors.
}
\label{fig:error_pos_step}
\end{figure}

We define two evaluation metrics used in the experiments,
positional error and step error, as defined in Algorithm \ref{alg1},
and shown in Figure \ref{fig:error_pos_step}.

Estimated segments should match actual segments,
regardless of the agent models, in terms of segmentation accuracy.
Therefore, we manually specified ground truth ``segmentation points,''
where the trajectory is segmented at these points,
for each of the training and test trajectories.
Then, we converted the segmentation result $s$ of trajectory $\y$
into the detection result $d = \{d_0, \ldots\}$ of segmentation points;
a sequence of Boolean values of the same length as $s$
and element $d_i$ is true (i.e., a segmentation point) if $s_i \neq s_{i-1}$;
otherwise, it is false.

We evaluated the detection results of segmentation points spatially and temporally.
The positional error counts the difference in L2 norm in two-dimensional space
between segmentation points in the ground truth and segmentation results.
The step error counts the time step difference (or index of the sequences).
Since we do not know which segmentation points correspond to those in other trajectories,
we chose the closest segmentation point for computing errors.
To prevent trivial results that minimize these errors
(for example, all the points are detected as segmentation points),
we added errors by switching the estimated and ground truth sequences.

\begin{algorithm}[t]

\KwIn{input trajectory $\y$, result $d$, ground truth $g$}

\def\pos{E_\mathrm{pos}}
\def\stp{E_\mathrm{step}}

\KwOut{$\pos$, $\stp$}

\SetKwProg{Fn}{Function}{:}{}
\SetKwFunction{CalcError}{CalcError}

\Fn{\CalcError{$\y, d, g$}}{

$\pos = \stp = 0$

\For{i}{
\If{$d_i$}{
$\hat{j} = \argmin_{g_j == True} |j - i|$

$\pos \mathrel{+}= \|\y_{\hat{j}} - \y_i\| $

$\stp \mathrel{+}= |\hat{j} - i|$
}
}
return $\pos$, $\stp$
}

$\pos, \stp = \CalcError{$\y, d, g$} + \CalcError{$\y, g, d$}$

$\pos, \stp \mathrel{/}= (N_\mathrm{est} + N_\mathrm{gt})$.

\caption{Calculation of positional and step errors.
Elements of $d$ and $g$ are assumed to be Boolean
($d_i$ or $g_i$ is true, if $\y_i$ is a (ground truth or estimated) segmentation point).
$N_{est}$ and $N_{gt}$ are the numbers of
estimated and ground truth segmentation points in a trajectory, respectively.}
\label{alg1}
\end{algorithm}

\subsection{Agent estimation with the improved MDA}

\begin{figure}[t]
\centering
\includegraphics[width=.6\linewidth]{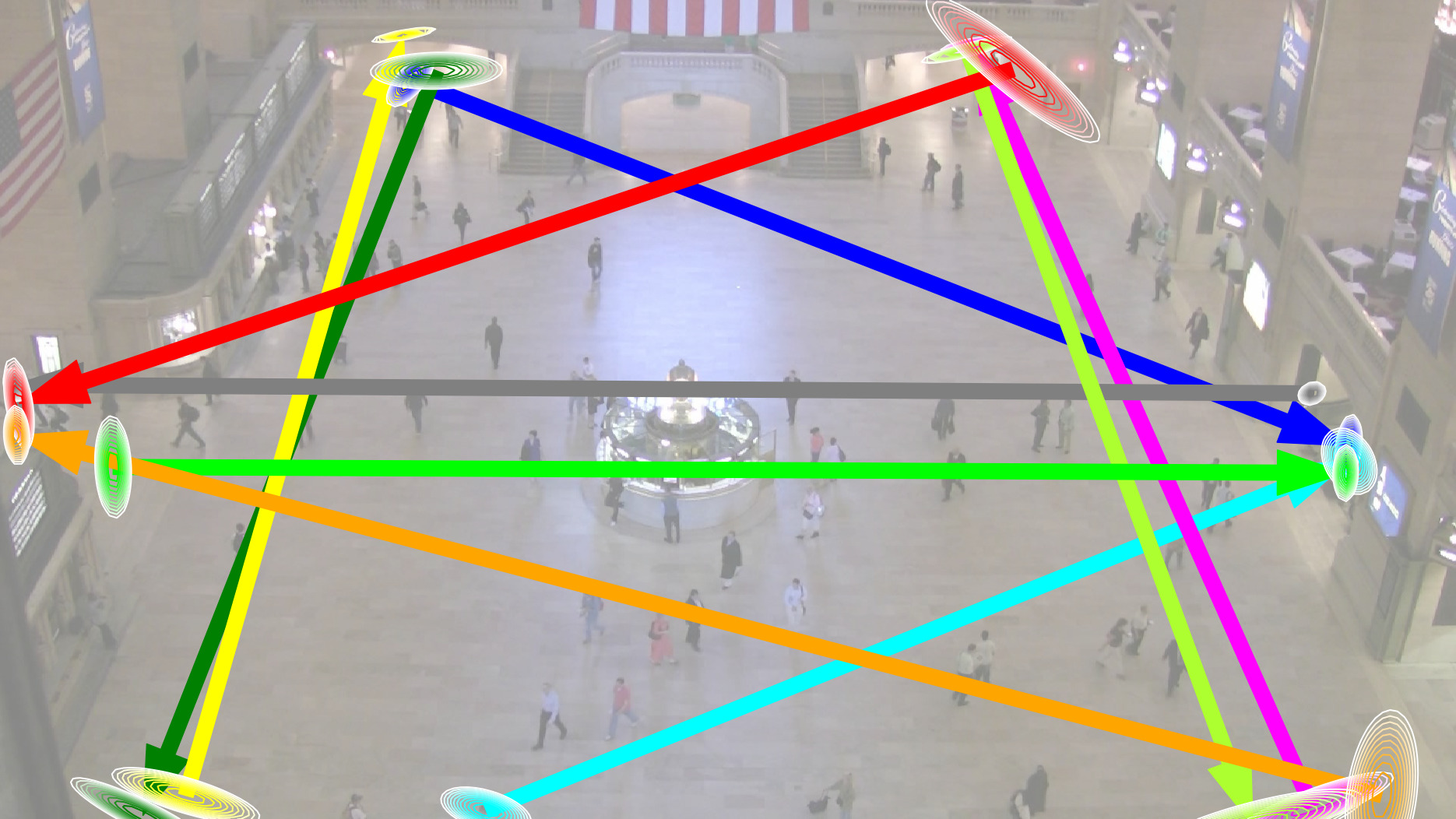}
\caption{Ten estimated agents estimated by the improved MDA.}
\label{fig:agents10}
\end{figure}

We propose iMDA because of the convergence problem that the original MDA suffers.
Here, we compare the convergence of the EM algorithm using a subset of
the Pedestrian Walking Path Dataset \cite{yi2015}.
First, we selected 1874 trajectories corresponding to approximately 10 agents,
each of which corresponds to a behavior connecting two exits from the scene.
Then, we trained the proposed iMDA with $M=10$ agents.
Figure \ref{fig:agents10} shows the estimated 10 agents.
Different agents are shown with arrows in different colors;
the arrows connect the start and end locations $\mu_s$ and $\mu_e$,
which are represented by Gaussian ellipses $p(\x_s)$ and $p(\x_e)$.
Locations $\mu_s$ and $\mu_e$ were initialized by k-means clustering
of the first and last points of the trajectories.

\begin{figure}[tp]
\centering
\includegraphics[width=.8\linewidth]{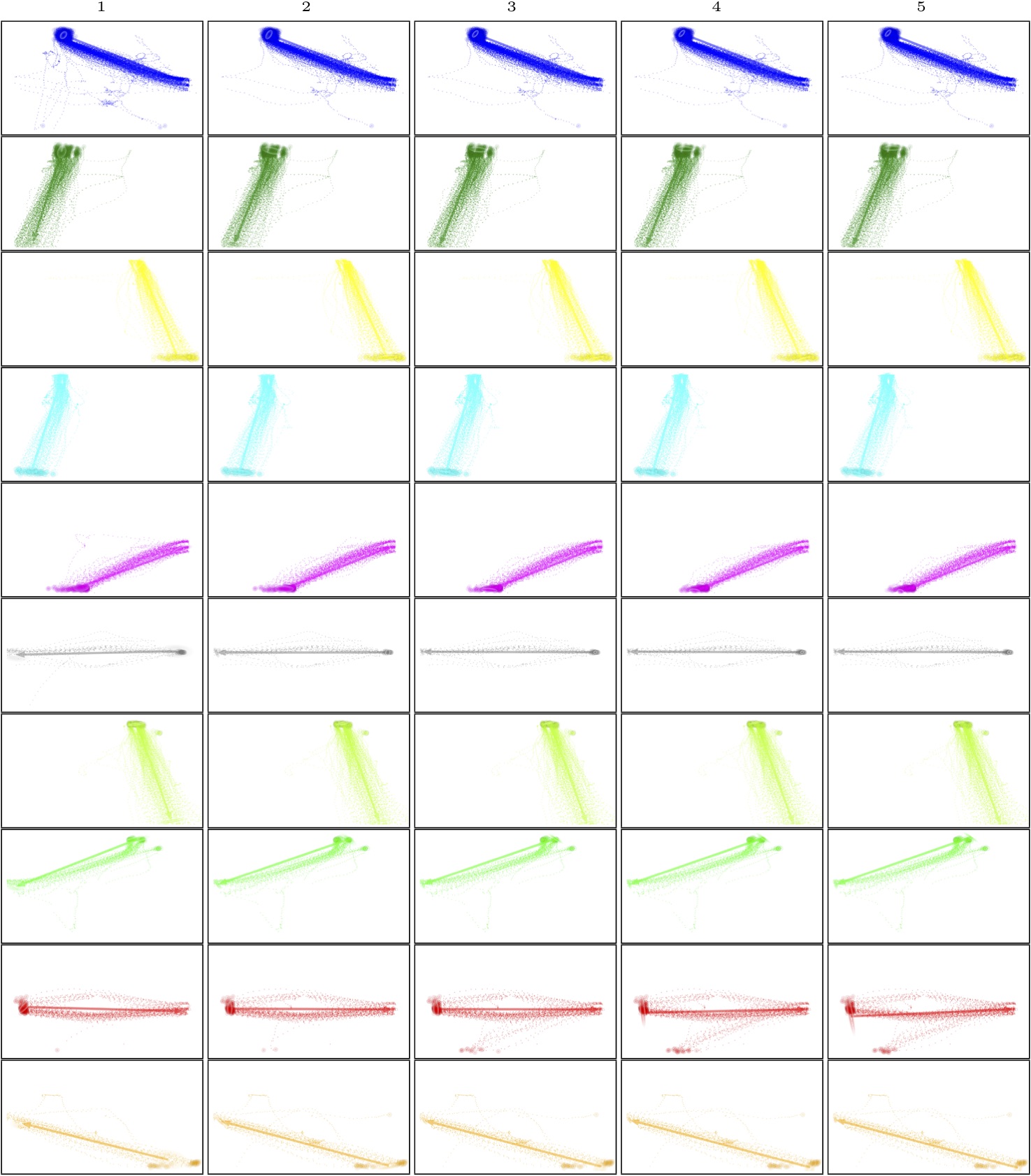}
\caption{Clustering results produced by the improved MDA. Each row shows clusters of trajectories
classified into agents. From left to right, the columns show the results at each iteration of the
estimation maximization algorithm.}
\label{fig:cluster-imda}
\end{figure}

\begin{figure}[tp]
\centering
\includegraphics[width=.8\linewidth]{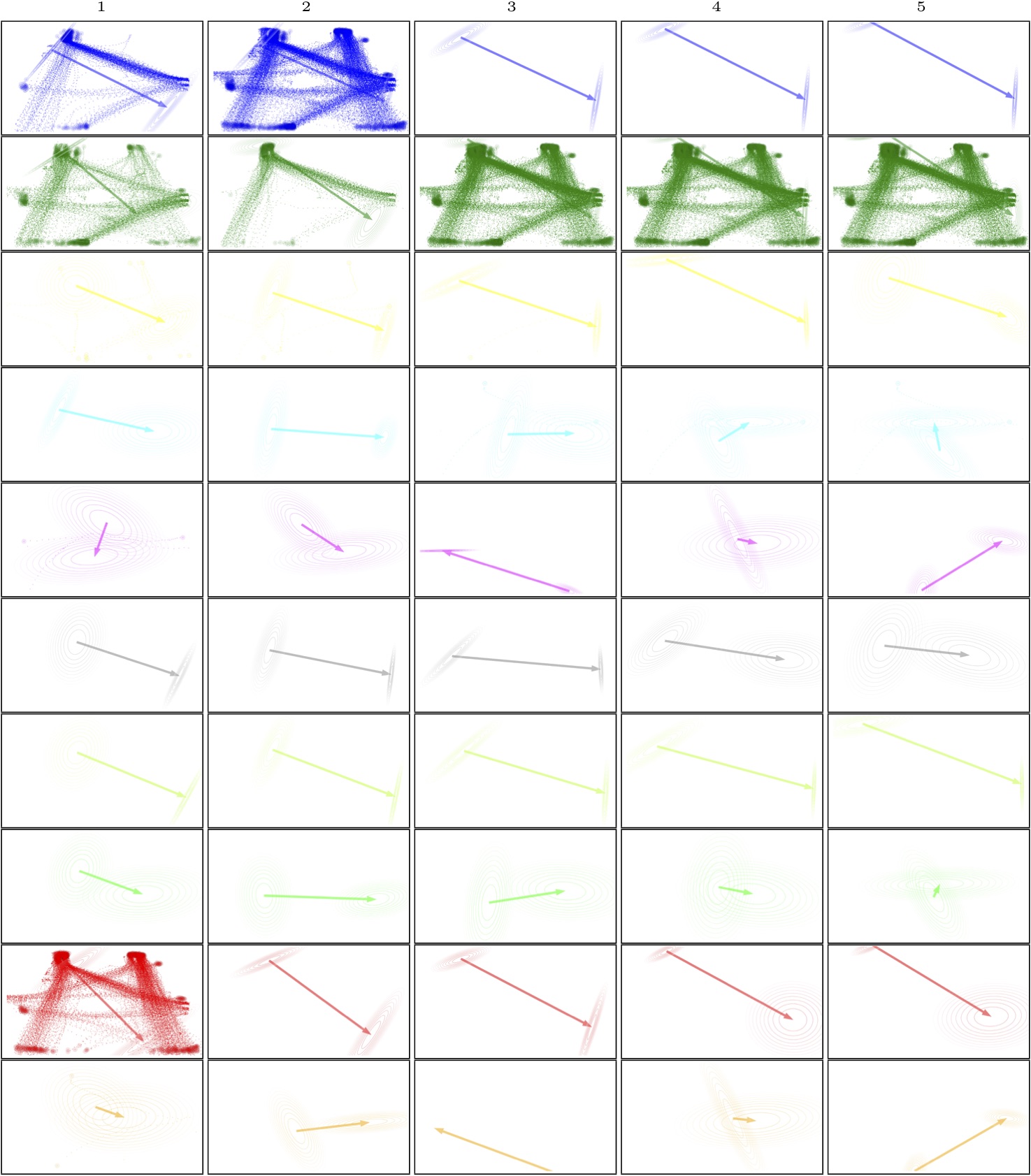}
\caption{Clustering results produced by the original MDA.}
\label{fig:cluster-mda}
\end{figure}

To obtain the agent models, trajectories were clustered into agents.
This clustering changes over iterations of the EM algorithm, which is
shown in Figure \ref{fig:cluster-imda}.
From left to right in the figure,
we can see that clustering converges after a small number of iterations,
while a few trajectories move from one cluster to another.
Figure \ref{fig:cluster-mda} shows the results of the original MDA.
Because of its instability, most trajectories go to a single cluster,
even when the initialization is the same as that of iMDA.
This clearly shows the effectiveness of the proposed method.

\begin{figure}[tp]
\centering
\includegraphics[width=.8\linewidth]{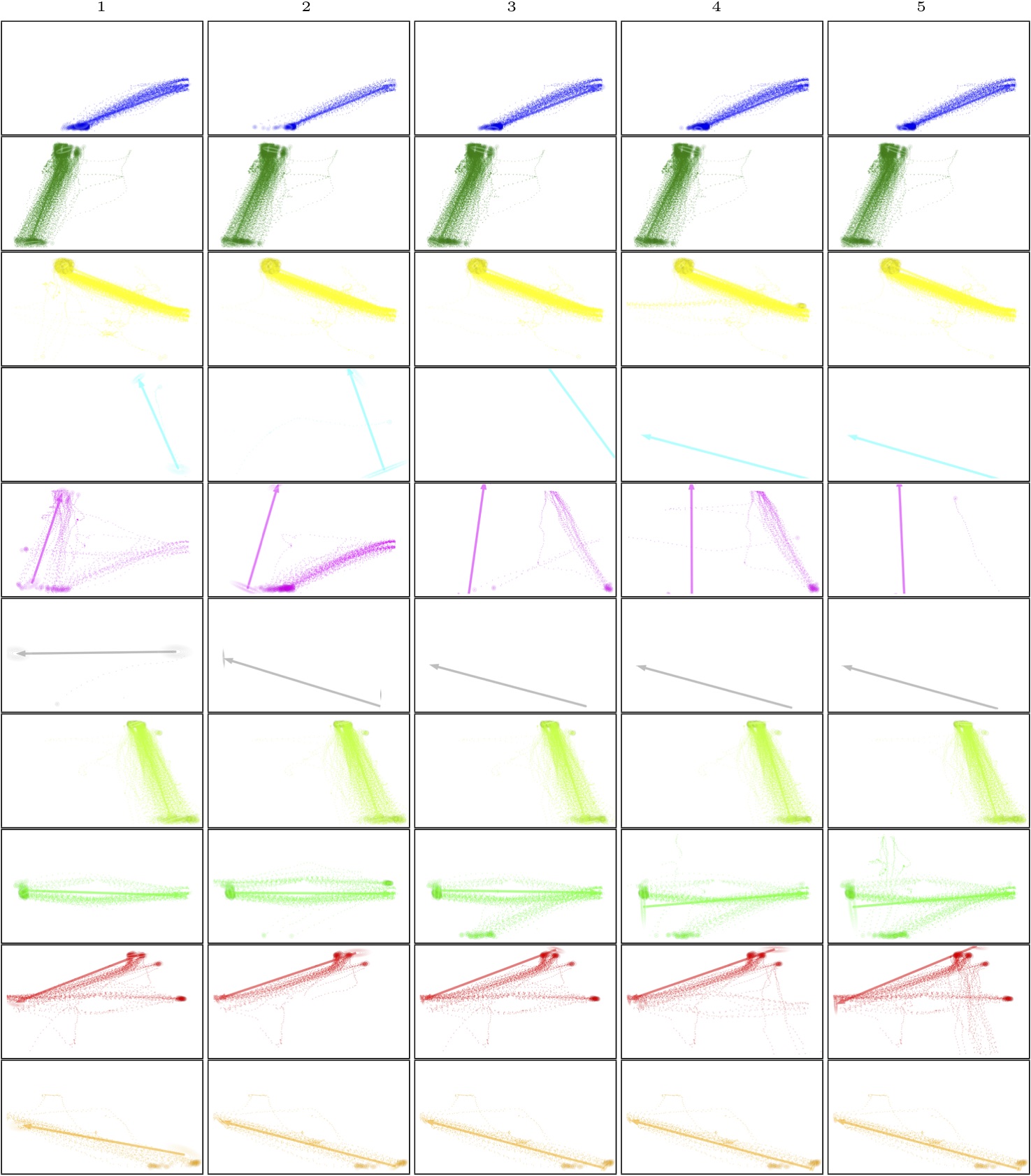}
\caption{Clustering results produced by the improved MDA without Poission distributions $p(t_s)$ and $p(t_e)$.}
\label{fig:cluster-imda-nopoisson}
\end{figure}

\begin{figure}[tp]
\centering
\includegraphics[width=.8\linewidth]{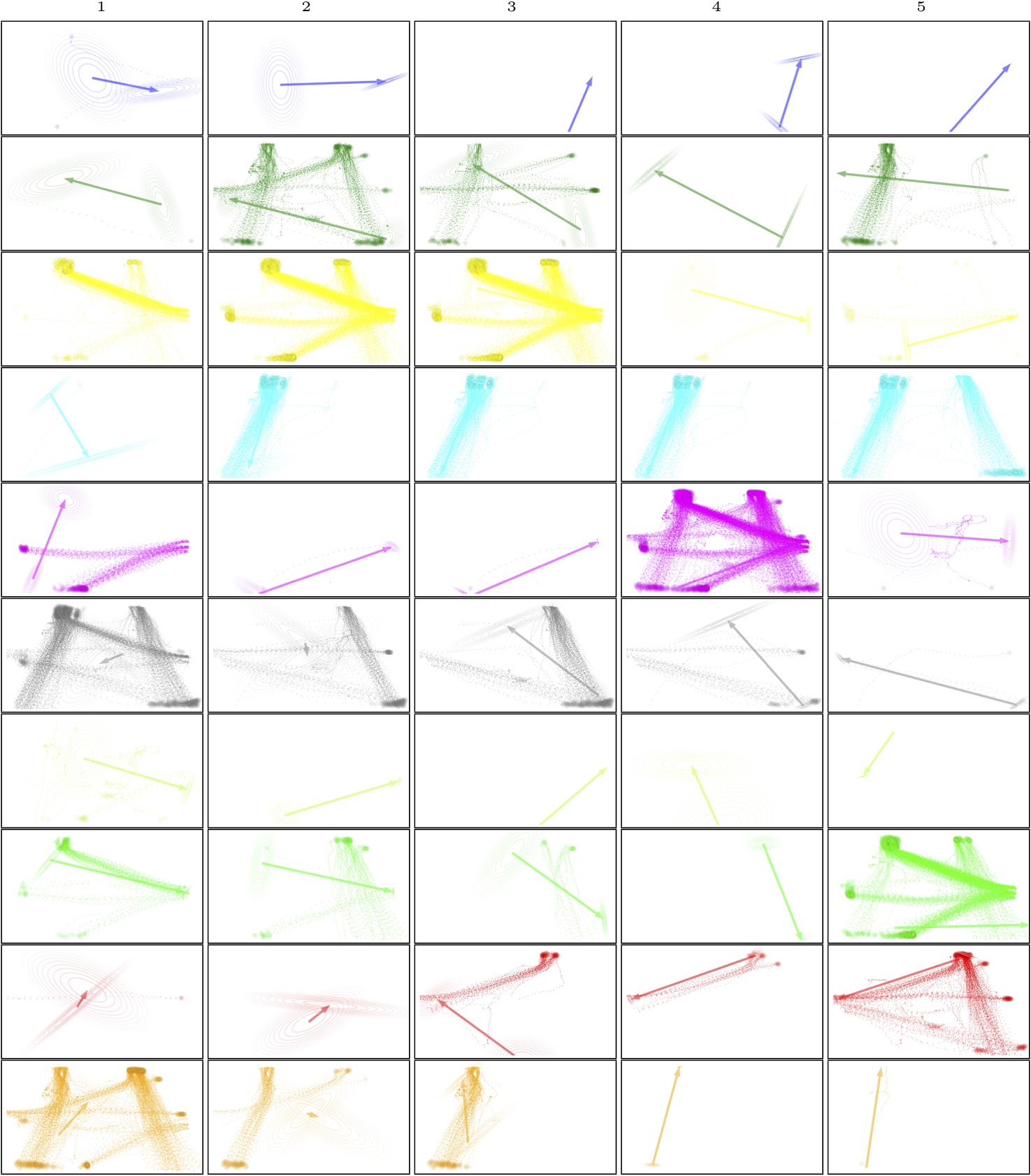}
\caption{Clustering results produced by the improved MDA without Gaussian distributions $p(\x_s)$ and $p(\x_e)$.}
\label{fig:cluster-imda-nogauss}
\end{figure}

The proposed iMDA has two factors in its E-step:
Gaussian distributions $p(\x_s)$ and $p(\x_e)$
and Poission distributions $p(t_s)$ and $p(t_e)$,
in weights $\gamma^k$ of the log likelihood of $Q(\Theta, \hat\Theta)$.
To observe the effect of these two factors on the results,
we omitted one or the other of the factors.
Figure \ref{fig:cluster-imda-nopoisson} shows the results
without Poission distributions,
and
Figure \ref{fig:cluster-imda-nogauss} the results
without Gaussian distributions.
In both figures, the clustering results are still unstable,
and both factors are necessary for improving the stability of
clustering trajectories and estimating agent models.

\clearpage

\subsection{Real data}
\label{subsec:actualdata}

To evaluate the methods using a real dataset, we manually annotated the 1874 trajectories used in the experiment by specifying points in the trajectories in which the destinations of the trajectories appeared to change.
Then, we performed 10-fold cross validation on the 1874 trajectories.
For each fold of the 10-fold cross validation, we performed
the iMDA model estimation and HMM training
with a different number of agents (between 9 and 12) on the training set.
The test set was used for segmentation and evaluation.
The results in Table \ref{table:comparison} (row ''iMDA+HMM'') show the averages and standard deviations for the 10-fold cross validation.
For the RDP method,
we also performed 10-fold cross validation. For each fold,
the best parameter $\epsilon$ (in terms of positional or step errors) was estimated for the training set,
and the estimated parameter was used for the test set.
The results in Table \ref{table:comparison} (row ''RDP'') show the averages and standard deviations for the 10-fold cross validation. The best $\epsilon$ selected using the cross validation is shown in the second column.

The errors of the proposed method and RDP are comparable:
Both the proposed method and RDP have positional errors of approximately 20 pixels.
However, RDP does not provide any semantic information for the segmentation.
In contrast, the proposed method can divide trajectories
into semantically meaningful segments by using the associated agent models,
which facilitates the understanding of pedestrians' behavior in a real-world scene.
The row ''MDA+HMM'' of Table \ref{table:comparison} shows the results when the original MDA is used for agent model estimation. 
This shows that the proposed improved MDA model performs better at segmentation. Furthermore, the results show that iMDA+HMM consistently outperforms MDA+HMM.

\begin{figure}[tp]
\centering

\begin{flushleft}
\subfigure[]{\includegraphics[width=0.2\linewidth]{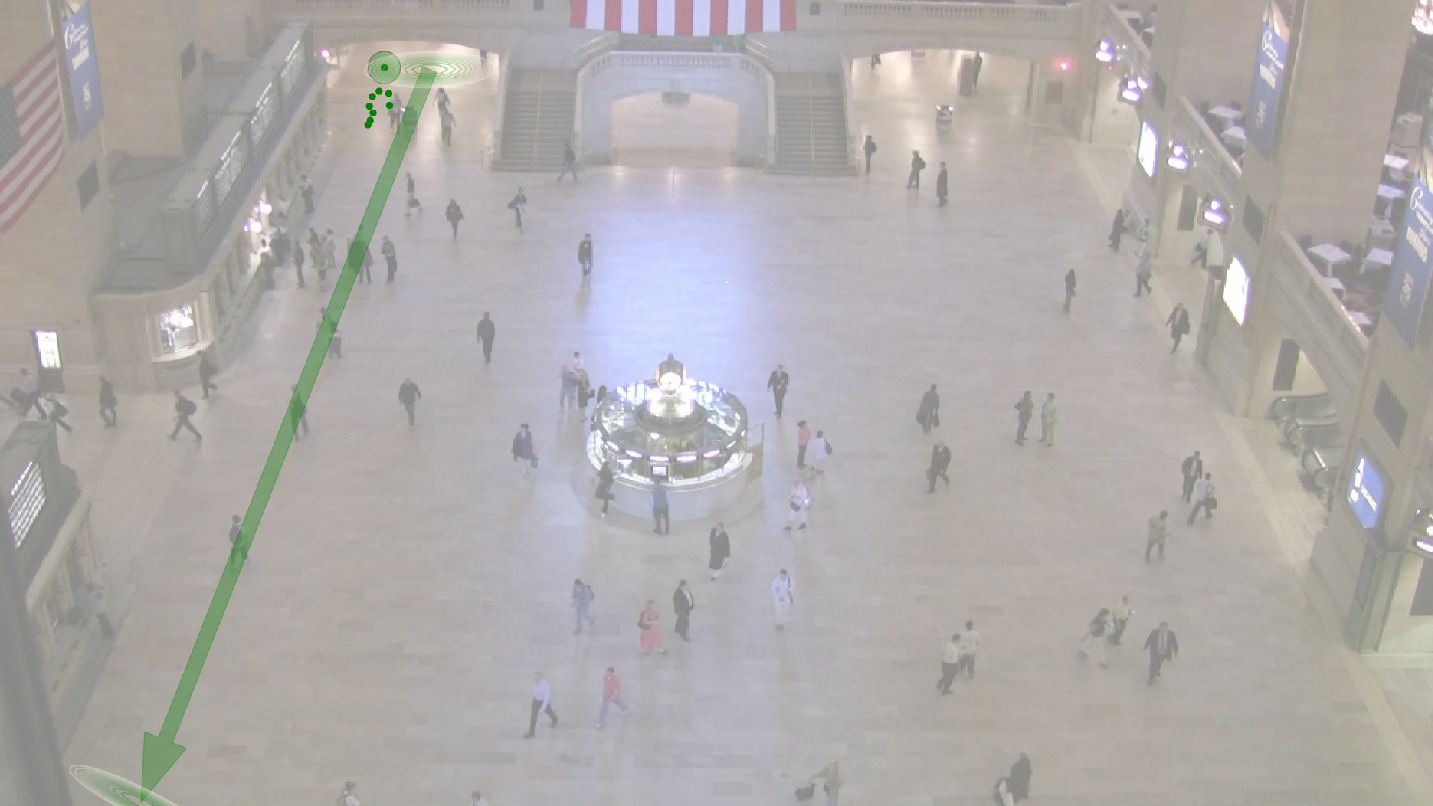}\label{fig:ped1-1}}%
\subfigure[]{\includegraphics[width=0.2\linewidth]{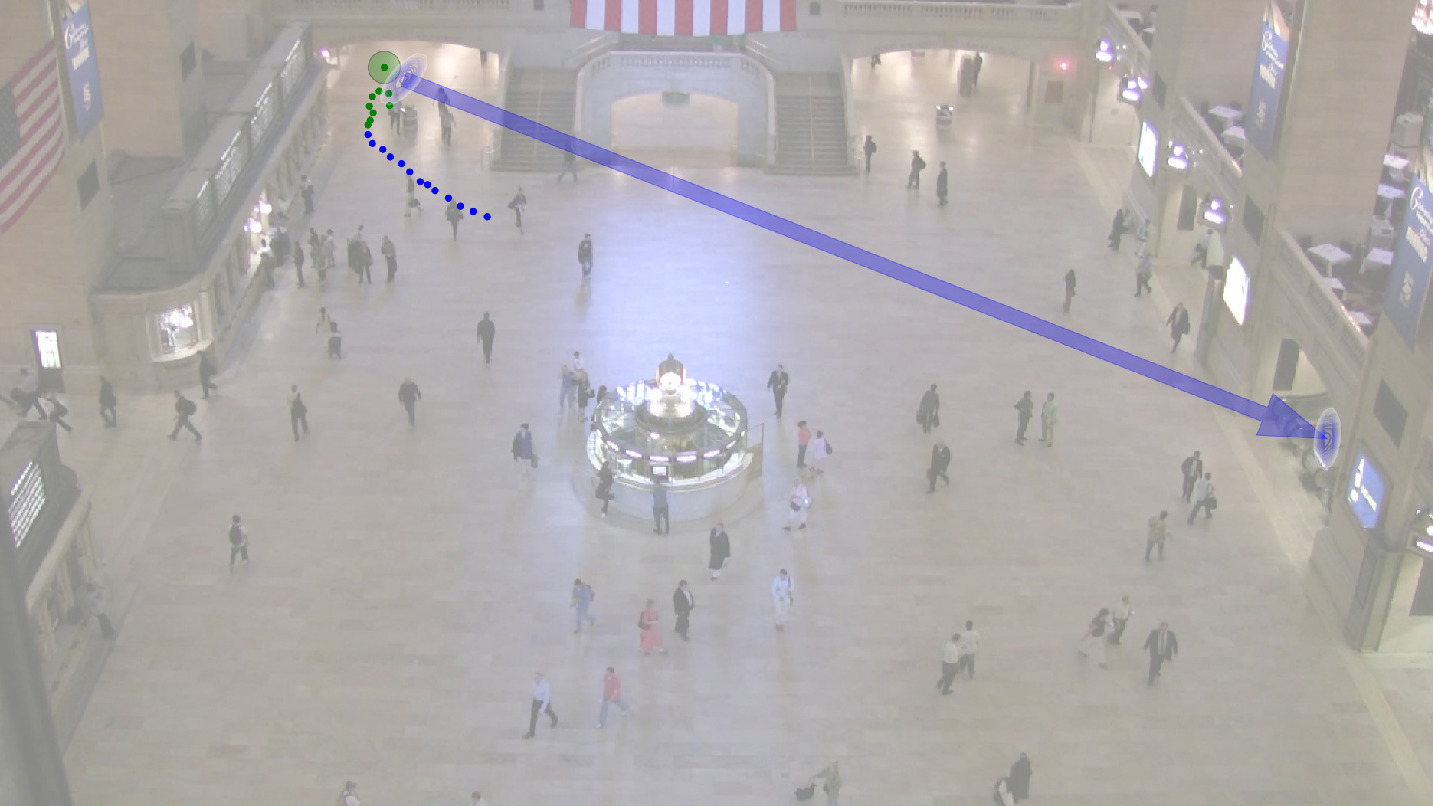}\label{fig:ped1-2}}%
\subfigure[]{\includegraphics[width=0.2\linewidth]{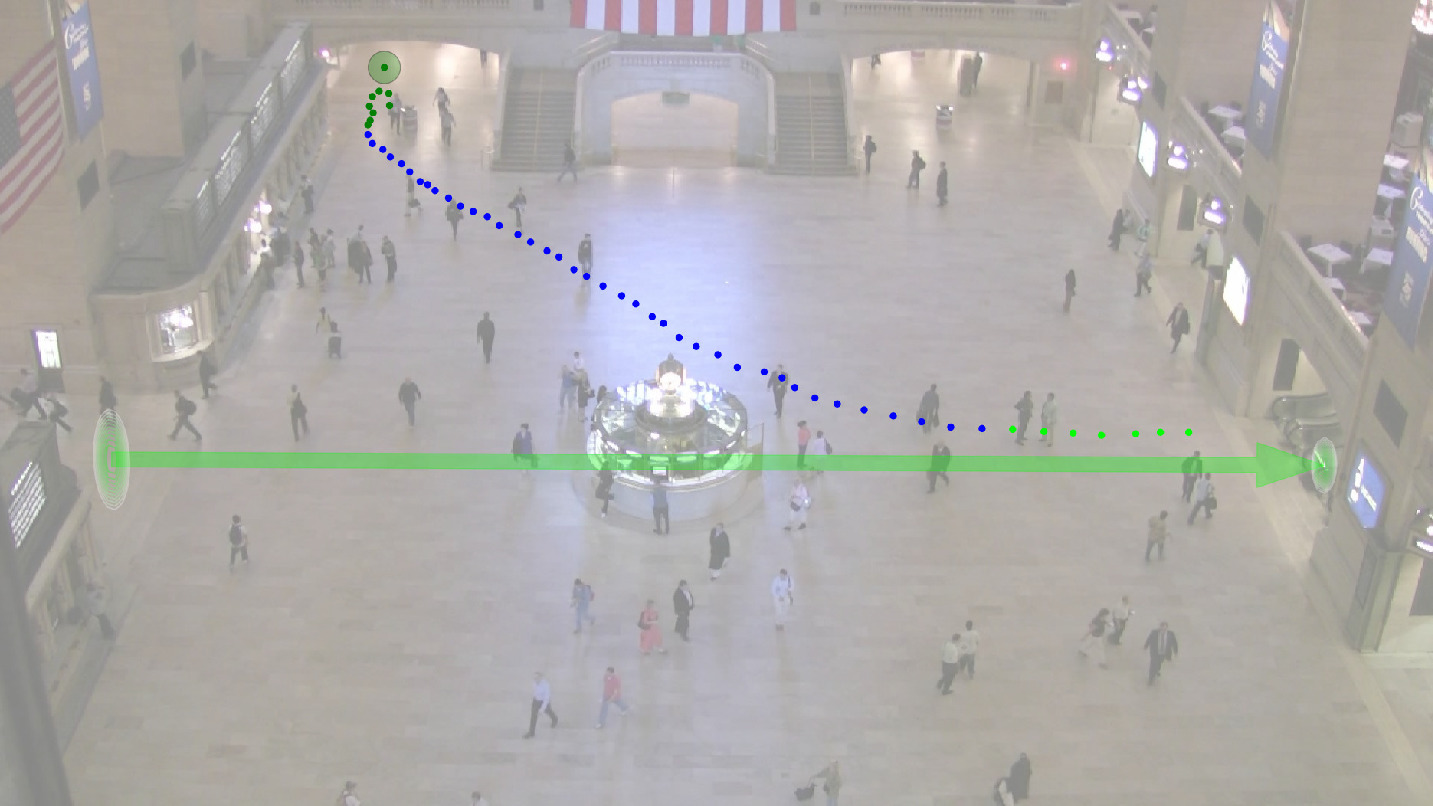}\label{fig:ped1-3}}%

\subfigure[]{\includegraphics[width=0.2\linewidth]{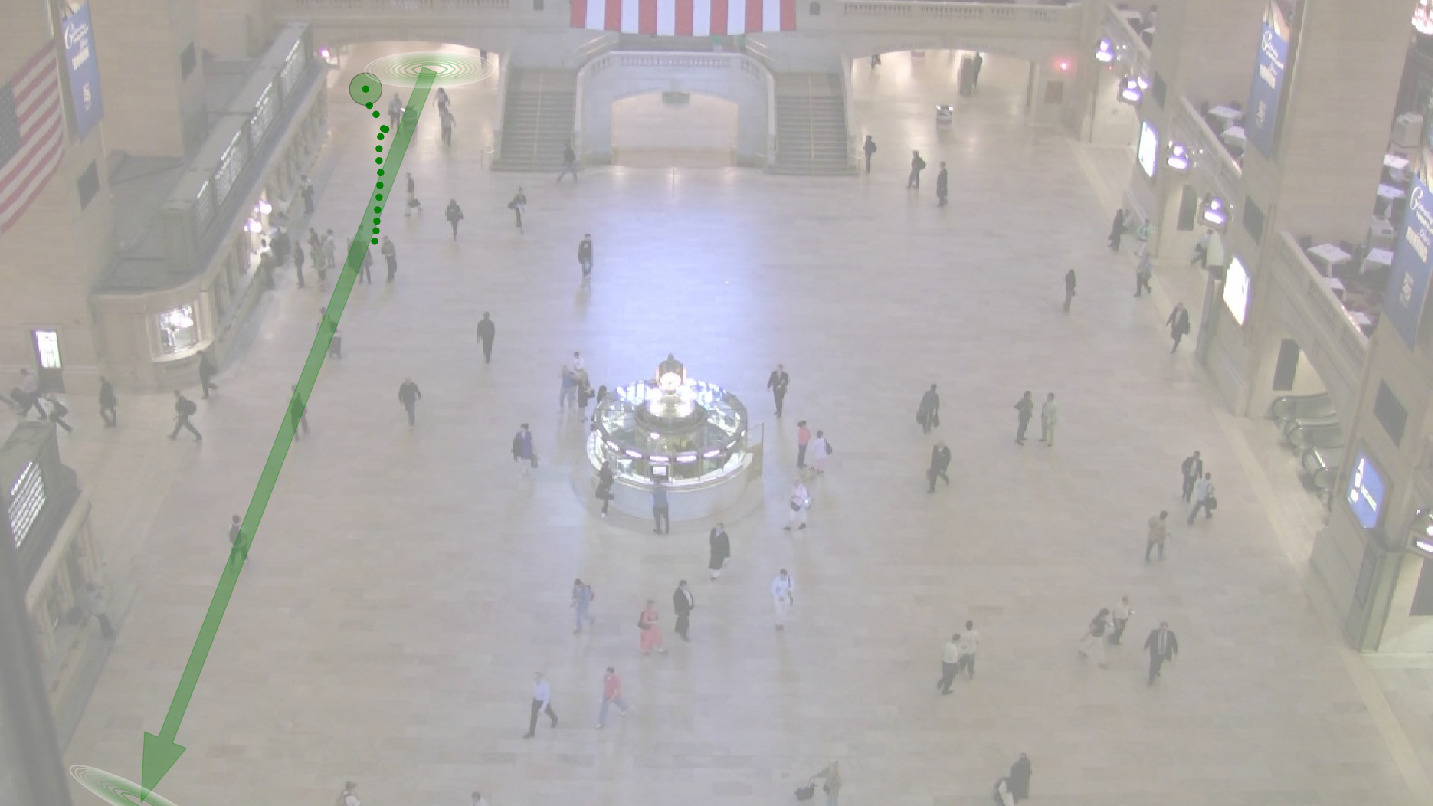}\label{fig:ped5-1}}%
\subfigure[]{\includegraphics[width=0.2\linewidth]{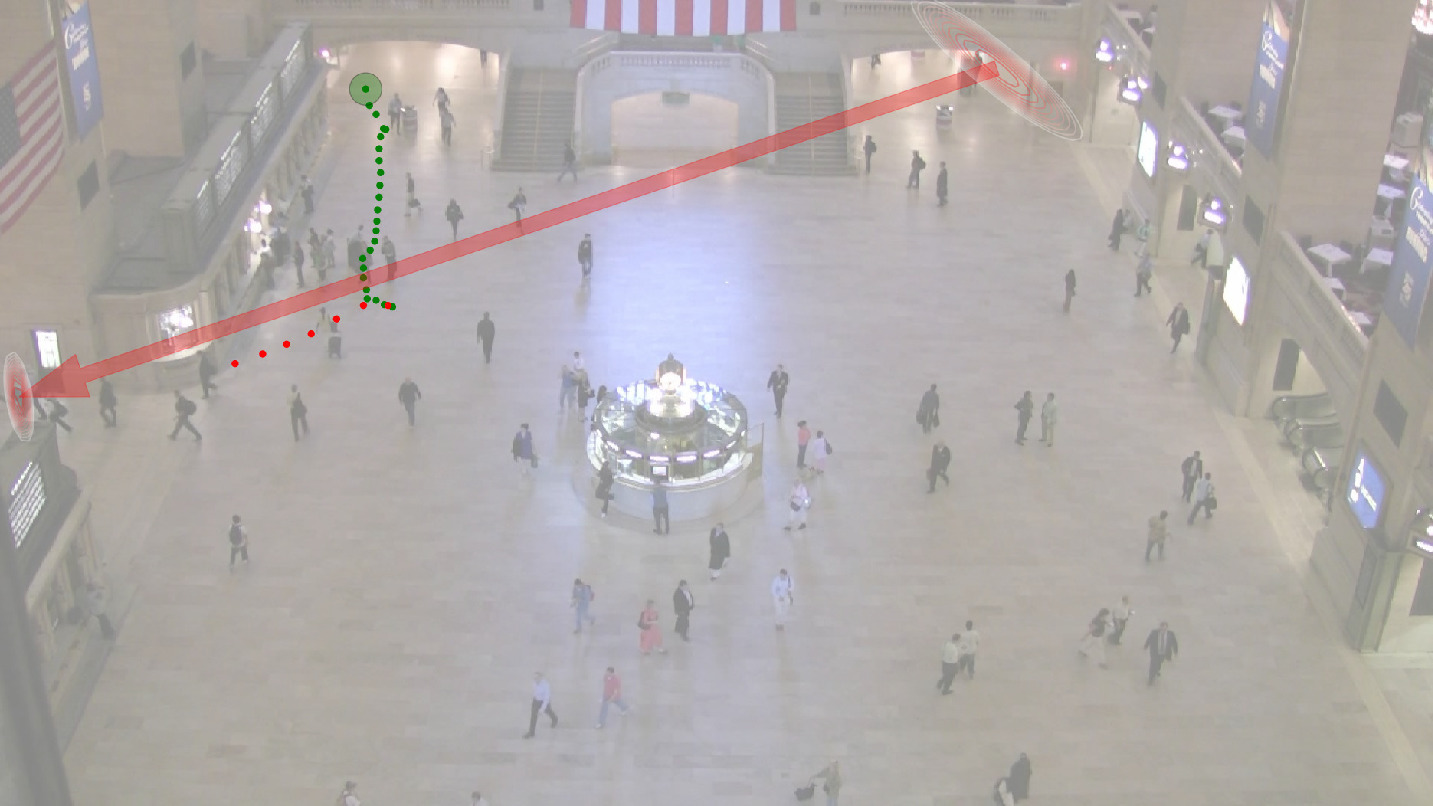}\label{fig:ped5-2}}%
\subfigure[]{\includegraphics[width=0.2\linewidth]{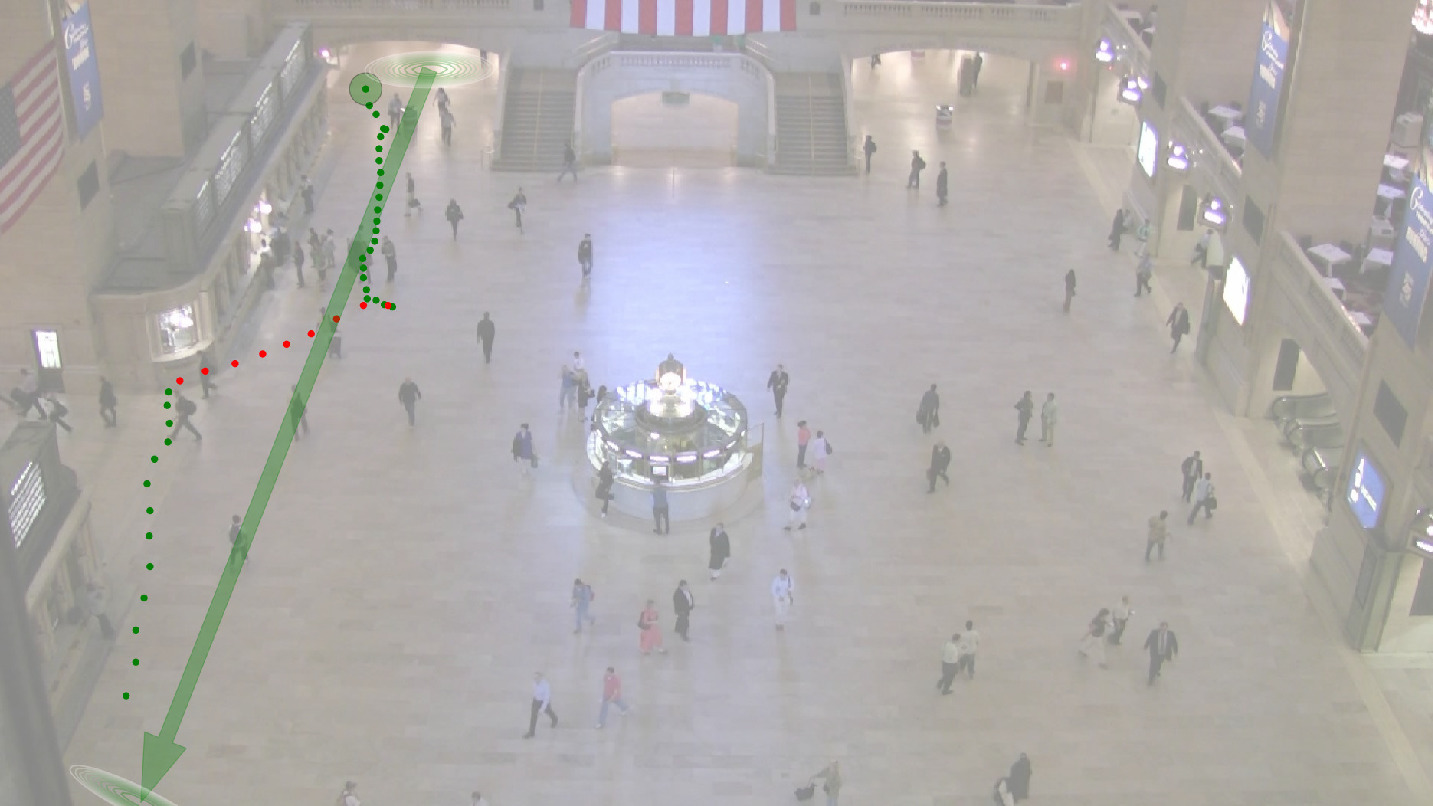}\label{fig:ped5-3}}%

\subfigure[]{\includegraphics[width=0.2\linewidth]{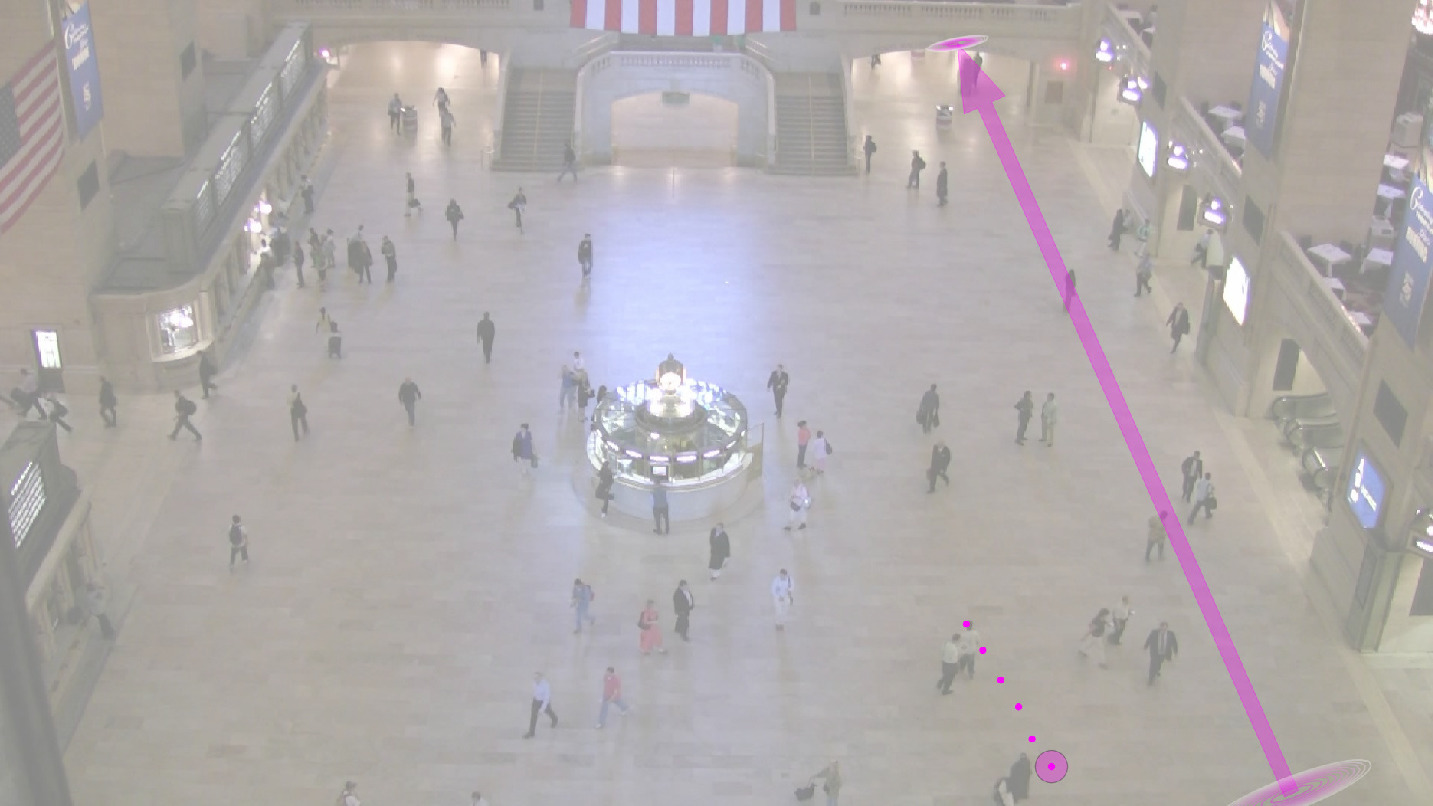}\label{fig:ped3-1}}%
\subfigure[]{\includegraphics[width=0.2\linewidth]{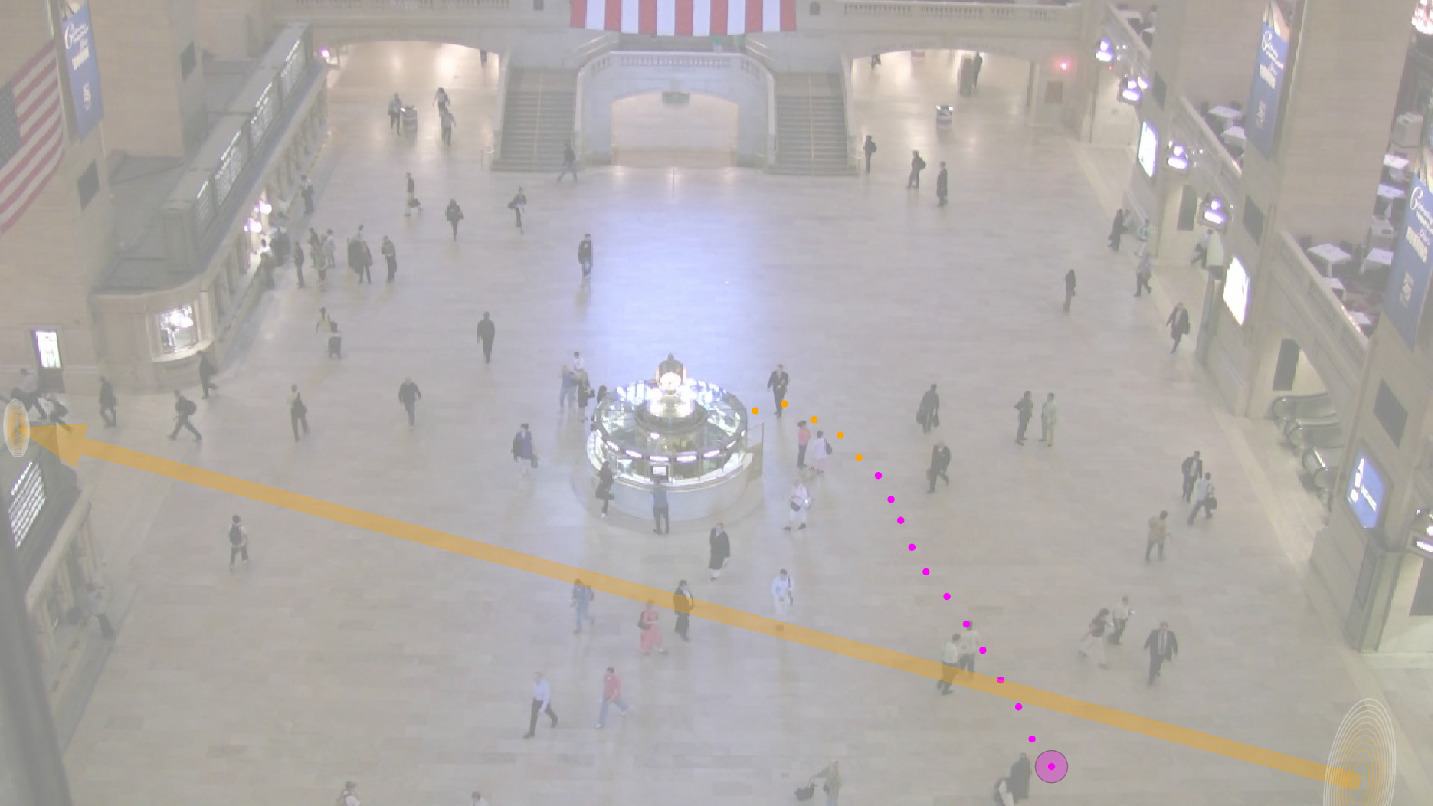}\label{fig:ped3-2}}%
\subfigure[]{\includegraphics[width=0.2\linewidth]{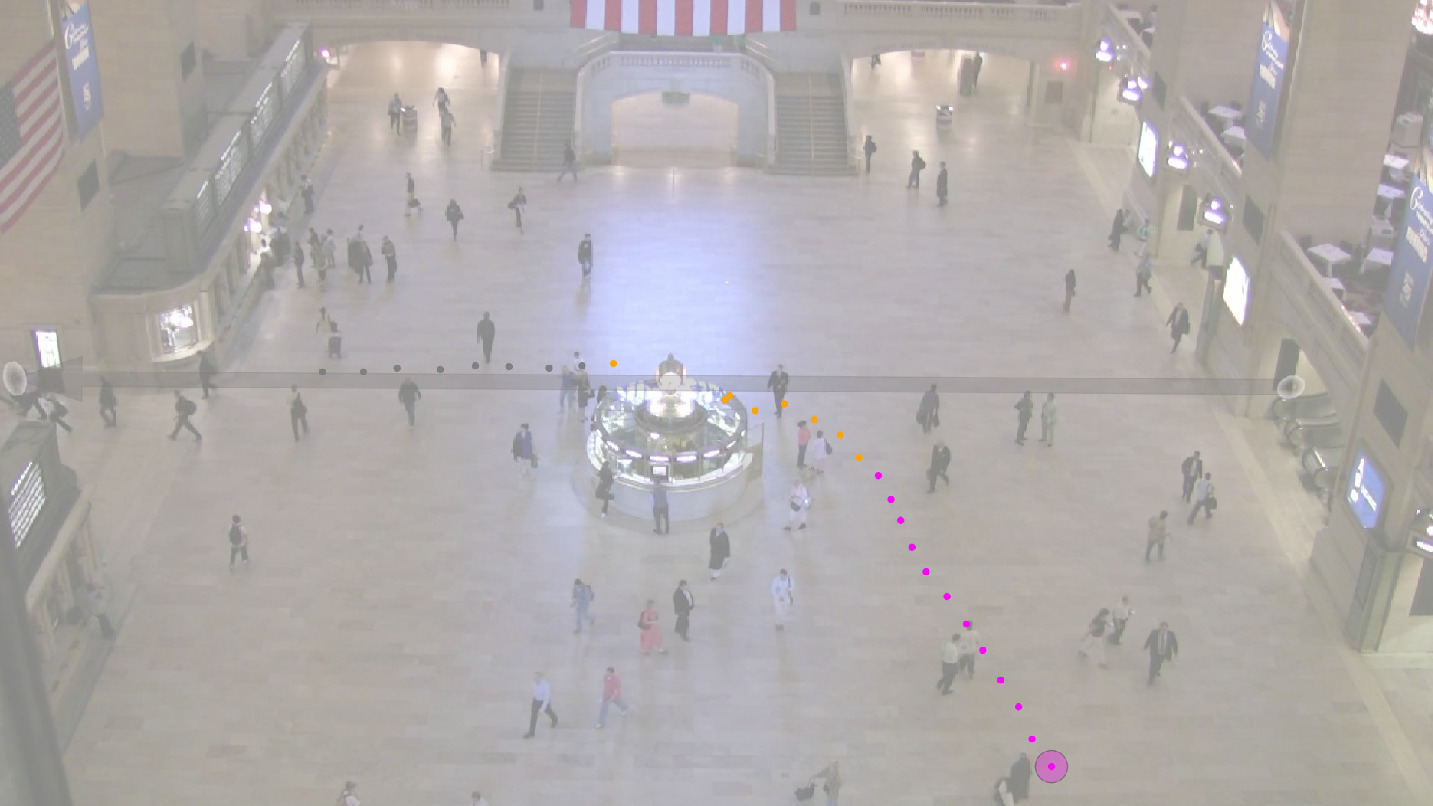}\label{fig:ped3-3}}%

\subfigure[]{\includegraphics[width=0.2\linewidth]{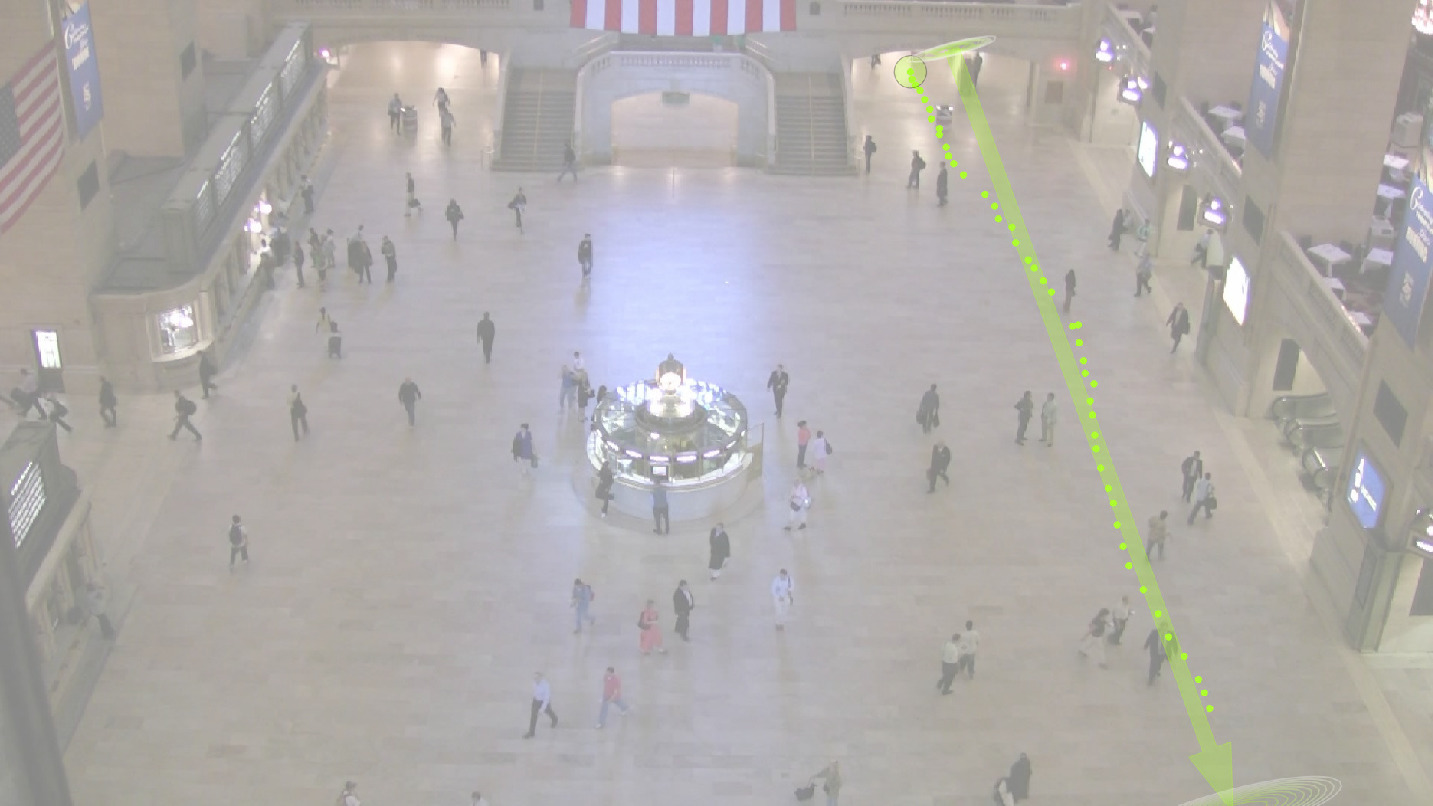}\label{fig:ped4-1}}%
\subfigure[]{\includegraphics[width=0.2\linewidth]{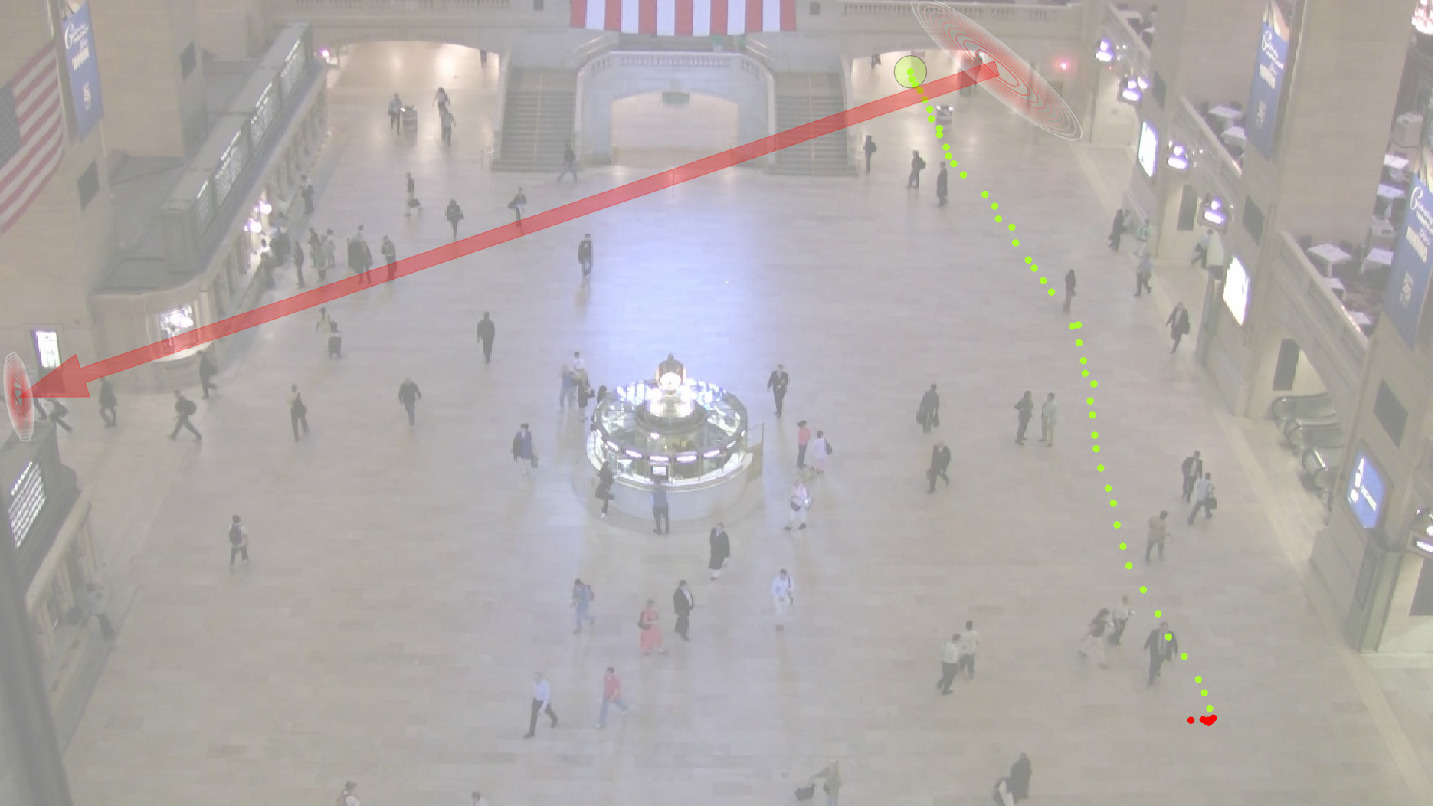}\label{fig:ped4-2}}%
\subfigure[]{\includegraphics[width=0.2\linewidth]{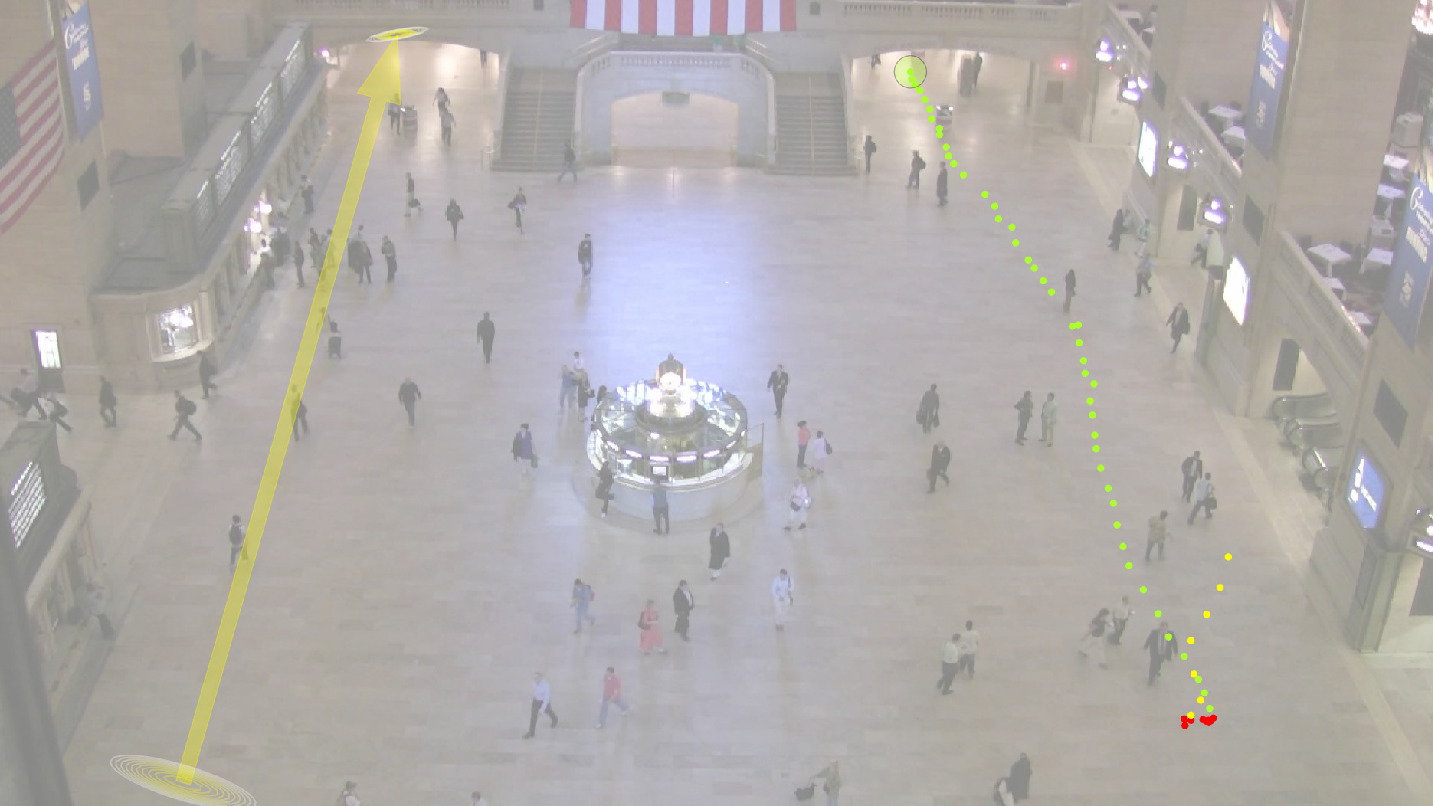}\label{fig:ped4-3}}%
\subfigure[]{\includegraphics[width=0.2\linewidth]{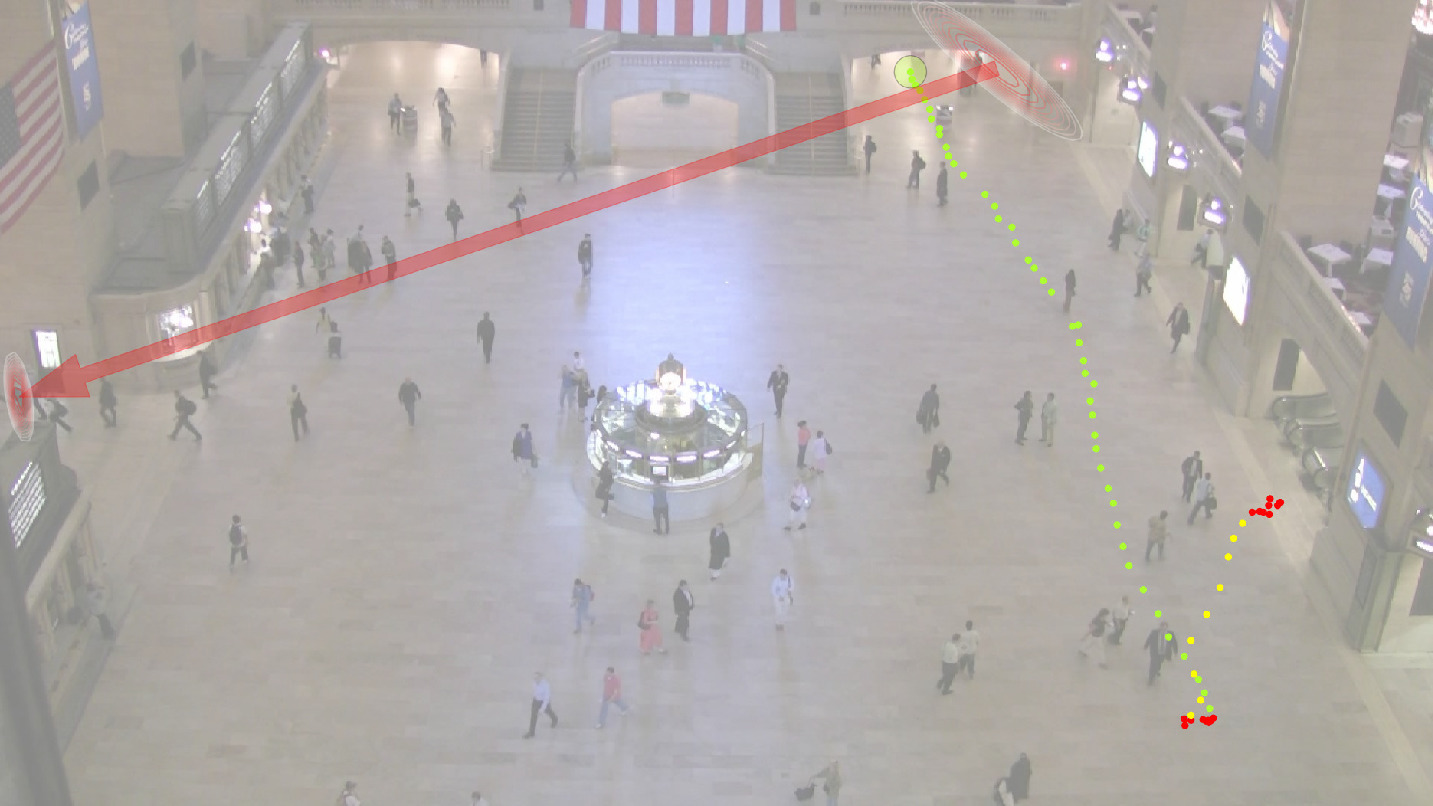}\label{fig:ped4-4}}%
\subfigure[]{\includegraphics[width=0.2\linewidth]{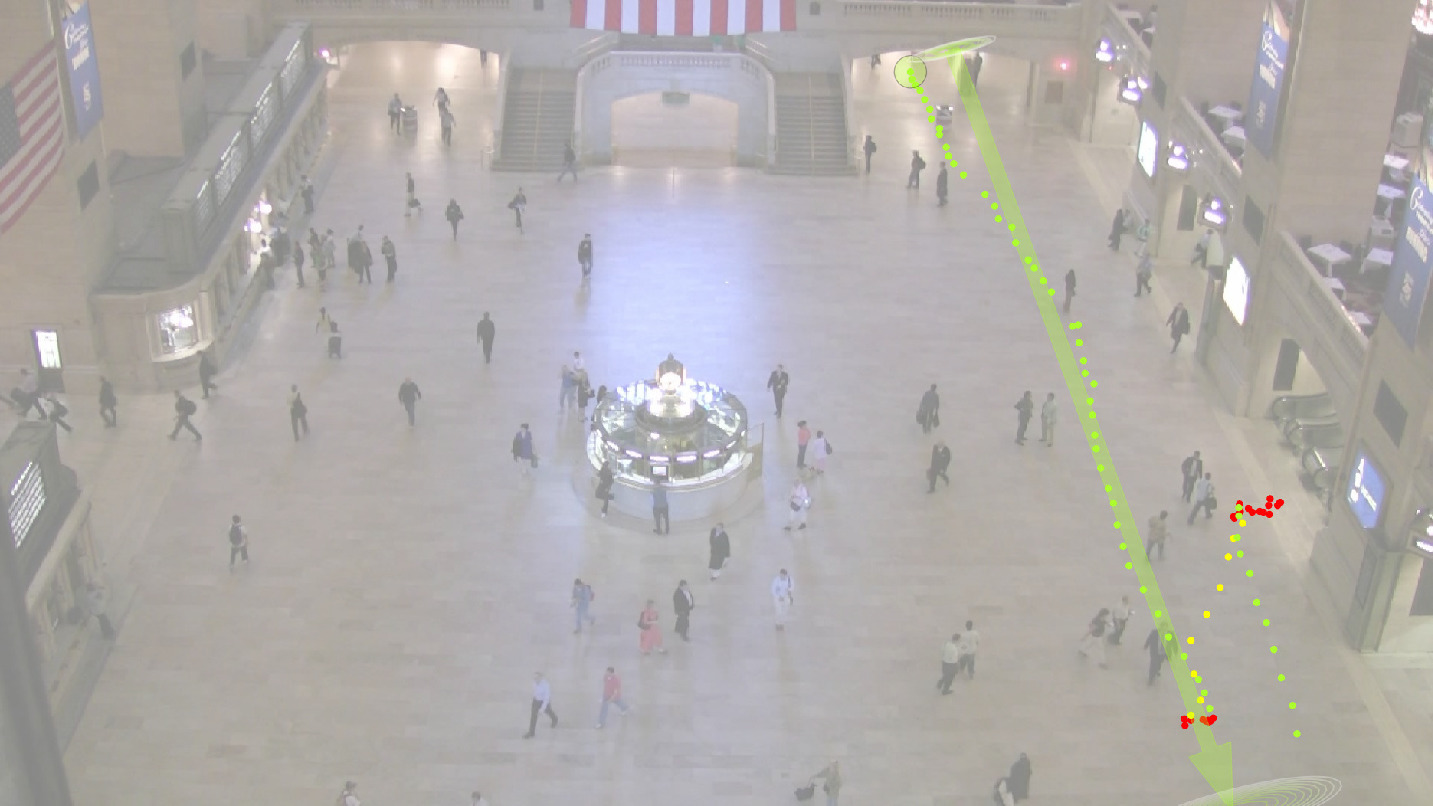}\label{fig:ped4-5}}%
\end{flushleft}

\caption{
Examples of segmentation results.
The arrow indicates the agent of the current position of the trajectory.
There are four trajectories:
(a--c), (d--f), (g--i), and (j--n).
The points of the trajectories are shown in the colors
of the corresponding agent models represented by arrows.
}
\label{fig:results}
\end{figure}

Figure \ref{fig:results} shows
four segmentation results.
In the first result (a--c),
the agents of the pedestrian are correctly visualized:
the pedestrian started from the top-left entrance,
first moved downward, and
then turned toward the exit at the right side.
Similar results were obtained
in the second (d--f) and third (g--i) trajectories.

Figure \ref{fig:results} (j--n) shows
the limitation of our approach.
The downward trajectory started from the right-top entrance
and turned its direction toward the right-top,
and then turned downward again.
We make two observations.
First, at the turning points, the agent from the right-top to the left side (shown in red in Figure \ref{fig:results} (k, m))
was estimated. This is due to the small step size of the trajectory movement,
which means the pedestrian may go nowhere, and an agent may be almost randomly assigned,
because no agent can describe the behavior. Procedures for rejecting such cases are needed.
Second,
when the pedestrian turned to the exit at the right side,
the agent from the bottom-left to top-right (shown in orange in Figure \ref{fig:results} (l))
was incorrectly estimated. Our HMM model uses agent dynamics only
and ignores belief parameters (start and end location, or the direction of agent arrow in the figure)
for computing the output probability matrix $\B$. Therefore, the selection of agents with similar dynamics suffers confusion. The incorporation of beliefs is left as future work.

\begin{table}[tb]
\caption{Experimental results for real data. Positional errors are in pixels.}
\label{table:comparison}

\begin{center}
\begin{tabular}{c|ccc}
Method & No. of Agent & Positional error & Step error \\
\hline
\multirow{5}{*}{MDA+HMM}
 &  9 & 42.93 $\pm$ 4.70 & 1.70 $\pm$ 0.22\\
 & 10 & 34.97 $\pm$ 6.71 & 1.40 $\pm$ 0.29\\
 & 11 & 43.49 $\pm$ 8.74 & 1.63 $\pm$ 0.27\\
 & 12 & 39.41 $\pm$ 6.27 & 1.52 $\pm$ 0.26\\
\hline
\multirow{5}{*}{iMDA+HMM (proposed)}
 &  9 & 22.32 $\pm$ 2.12 & 0.91 $\pm$ 0.13 \\
 & 10 & 22.32 $\pm$ 2.51 & 0.92 $\pm$ 0.13 \\
 & 11 & 22.88 $\pm$ 2.31 & 0.94 $\pm$ 0.12 \\
 & 12 & 22.38 $\pm$ 2.83 & 0.91 $\pm$ 0.13 \\
\hline
RDP (best positional error)
& $\epsilon=137.3 \pm 10.8$ & 21.93 $\pm$ 3.23 & 
\\
RDP (best step error)
& $\epsilon=108.7 \pm \phantom{1}2.7$ & 
& 0.95 $\pm$ 0.14\\
\hline
\end{tabular}

\end{center}

\end{table}

\subsection{Behavior analysis}

In the previous section, we provided a quantitative performance analysis and showed the segmentation results. In this section, we provide a qualitative behavior analysis to demonstrate the effectiveness of the proposed semantic segmentation method.

\subsubsection{Transition between agents}

\def\diag{\mathrm{diag}}
\def\normr{\mathrm{normr}}

\begin{figure}[t]
\centering
\includegraphics[width=.5\linewidth]{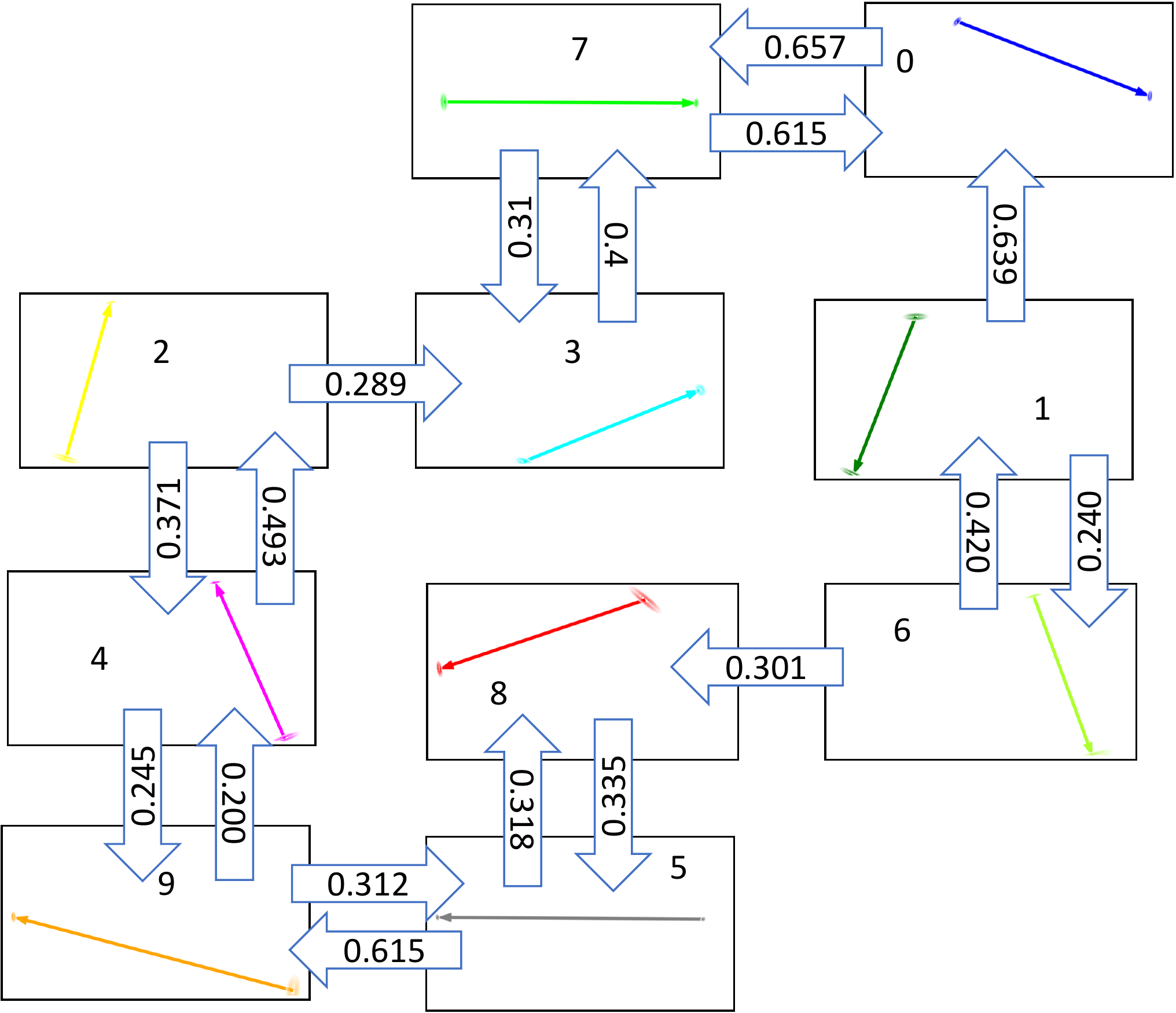}
\caption{Transition between agent models.
The arrows between rectangles represent the normalized transition probabilities.
The agent numbers shown near arrows are the same as in Figure \ref{fig:cluster-imda}.
}
\label{fig:normalized_transition_prob_0.2}
\end{figure}

Table \ref{tab:transition_prob} shows
the transition probability matrix $\A$ estimated by the HMM training
with 1874 trajectories.
Diagonal elements represent the probability that successive points
in a trajectory have the same agent model; it is reasonable that the probability is very close to 1.
To visualize the transition between agents more clearly,
we define a normalized transition probability matrix $\A'$:
\begin{equation}
\A' = \normr(\A - \diag(\diag(\A))),
\end{equation}
where $\normr$ normalizes the rows of a given matrix.
The normalized $\A'$ (shown in Table \ref{tab:normalized_transition_prob})
represents transitions to other agents (not including itself),
and therefore helps us to understand the relations between agents.
Figure \ref{fig:normalized_transition_prob_0.2} visualizes $\A'$
only with transitions having probabilities larger than 0.2.
This shows that agents tend to transit to each other
if they share start or end locations (similar beliefs)
or if their direction is similar (dynamics).

We can make two observations from this figure.
First, agents are switched if they share
the same start or goal point, which is reasonable.
Second, interestingly, some of these transitions are asymmetric; for instance,
agent 1 to 0, 2 to 3, and 6 to 8.
Agent 0 represents the top-left entrance and the escalator on the right-side,
and agent 1 represents the top-left entrance and the exit on the left-bottom.
Both agents 0 and 1 share the same starting point; however, the dominant transition
is from 1 to 0. 
Although some pedestrians who are following the flow to the bottom-left exit at first may turn to the escalator (agent 1 to 0), the
opposite rarely happens (agent 0 to 1). 
This observation might be useful, for example, for suggesting the placement of additional signboards to guide people from the top-left entrance to the right escalator more effectively.

\begin{table}
\caption{Transition matrix $\A$ obtained from 1874 trajectories.
Row $i$ represents probabilities of transition from agent $i$ to agent $j$.
The order of the agents is the same as in Figure \ref{fig:cluster-imda}.
}
\label{tab:transition_prob}
\centering
{\scriptsize
\begin{tabular}{l|llllllllll}
&0&1&2&3&4&5&6&7&8&9\\\hline
0&0.972 & 0.002 & 0.    & 0.006 &  0.    & 0.    & 0.002 & 0.019 & 0.    & 0.    \\
1&0.04  & 0.937 & 0.001 & 0.001 &  0.    & 0.    & 0.015 & 0.001 & 0.002 & 0.003 \\
2&0.    & 0.    & 0.974 & 0.008 &  0.01  & 0.    & 0.002 & 0.    & 0.003 & 0.003 \\
3&0.006 & 0.005 & 0.006 & 0.953 &  0.002 & 0.    & 0.003 & 0.019 & 0.004 & 0.001 \\
4&0.    & 0.002 & 0.01  & 0.002 &  0.979 & 0.001 & 0.    & 0.    & 0.    & 0.005 \\
5&0.    & 0.001 & 0.    & 0.001 &  0.002 & 0.931 & 0.    & 0.    & 0.022 & 0.043 \\
6&0.004 & 0.014 & 0.    & 0.001 &  0.    & 0.    & 0.968 & 0.    & 0.01  & 0.002 \\
7&0.058 & 0.    & 0.001 & 0.029 &  0.001 & 0.    & 0.    & 0.906 & 0.004 & 0.001 \\
8&0.001 & 0.008 & 0.005 & 0.002 &  0.    & 0.024 & 0.014 & 0.004 & 0.929 & 0.013 \\
9&0.009 & 0.014 & 0.012 & 0.01  &  0.032 & 0.051 & 0.021 & 0.005 & 0.007 & 0.838 \\
\end{tabular}
}

\end{table}

\begin{table}
\caption{Normalized transition matrix $\A'$ obtained from 1874 trajectories
}
\label{tab:normalized_transition_prob}
\centering
{\scriptsize
\begin{tabular}{l|llllllllll}
&0&1&2&3&4&5&6&7&8&9\\\hline
0 & 0.    & 0.059 & 0.001 & 0.196 &  0.    & 0.    & 0.073 & 0.657 & 0.002 & 0.012 \\
1 & 0.639 & 0.    & 0.011 & 0.009 &  0.006 & 0.    & 0.24  & 0.015 & 0.033 & 0.047 \\
2 & 0.001 & 0.009 & 0.    & 0.289 &  0.371 & 0.005 & 0.082 & 0.012 & 0.126 & 0.105 \\
3 & 0.132 & 0.11  & 0.122 & 0.    &  0.041 & 0.    & 0.069 & 0.403 & 0.095 & 0.028 \\
4 & 0.007 & 0.076 & 0.493 & 0.087 &  0.    & 0.07  & 0.02  & 0.003 & 0.    & 0.245 \\
5 & 0.    & 0.018 & 0.006 & 0.008 &  0.034 & 0.    & 0.    & 0.    & 0.318 & 0.615 \\
6 & 0.129 & 0.42  & 0.011 & 0.043 &  0.008 & 0.001 & 0.    & 0.012 & 0.301 & 0.075 \\
7 & 0.615 & 0.003 & 0.01  & 0.31  &  0.011 & 0.    & 0.    & 0.    & 0.041 & 0.009 \\
8 & 0.015 & 0.118 & 0.075 & 0.023 &  0.003 & 0.335 & 0.194 & 0.06  & 0.    & 0.176 \\
9 & 0.058 & 0.083 & 0.075 & 0.064 &  0.2   & 0.312 & 0.129 & 0.033 & 0.045 & 0.    \\
\end{tabular}
}

\end{table}

\subsubsection{Agent occurrence map}

\begin{figure}[t]
\centering

\includegraphics[width=.45\linewidth]{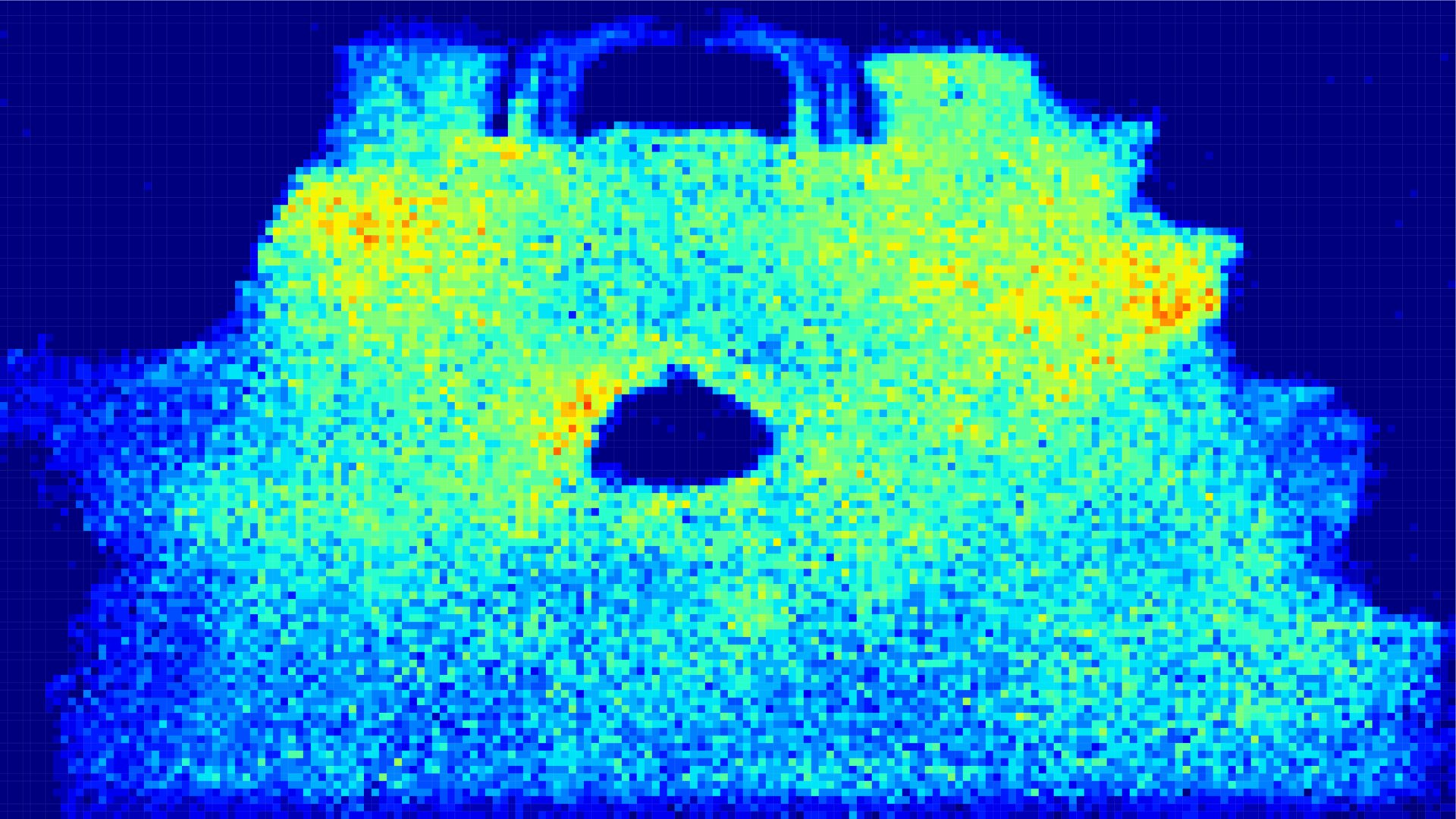}

\caption{
Agent occurrence map
showing the number of agents that appear in each of the 10$\times$10 blocks of the scene. The colors indicate that many agents pass through the red blocks, and fewer agents pass through the blue blocks.}
\label{fig:densemap_10x10_-crop}
\end{figure}

Figure \ref{fig:densemap_10x10_-crop} shows an agent occurrence map.
In this experiment, we used all 12684 trajectories in the dataset,
applied iMDA with 20 agents, performed HMM training, and then segmented all the trajectories.
Then, we counted the number of agents that appear in each of 10$\times$10 blocks of the scene of size 1920$\times$1080.
For example, a block has a count of 1 if all trajectory segments passing through the block
are assigned to the same agent.
In the figure, a block is shown in red if the trajectory segments of many different agents pass
through it.

There are mainly three areas with higher counts; the ticket counter just below the top-left entrance, the information booth at the center, and the right exit just above the right-side escalator.
This means that many agents appear in these areas. In other words, there are many 
people in these areas that are coming from and going to different locations. Therefore, these areas may be crowded, and the flow of pedestrians may not be smooth, as is the case for the right exit. The high number of agents in the area in front of the ticket counter may represent queues because people move slowly in many directions when standing in a queue. Furthermore, the left and right sides of the information booth are not symmetric, which might suggest an imbalance of activity on the left and right sides of the booth.

\subsubsection{Agent density maps of trajectory segments}

Here, we show how agents correspond to segments in the scene.
We used the segments of the 12684 trajectories obtained during the previous examination of the agent occurrence map. Extracting segments corresponding to a specific agent
may explain the behavior of the agent in terms of segments of trajectories, instead of the entire trajectories, as shown in Figure \ref{fig:cluster-imda}.
We plotted the density maps of points of segments using kernel density estimation (KDE)
because KDE is more effective for visually understanding distributions of segments
than for plotting a large number of points of segments.

\begin{figure}[tp]
\centering

\subfigure[]{\includegraphics[width=0.3\linewidth]{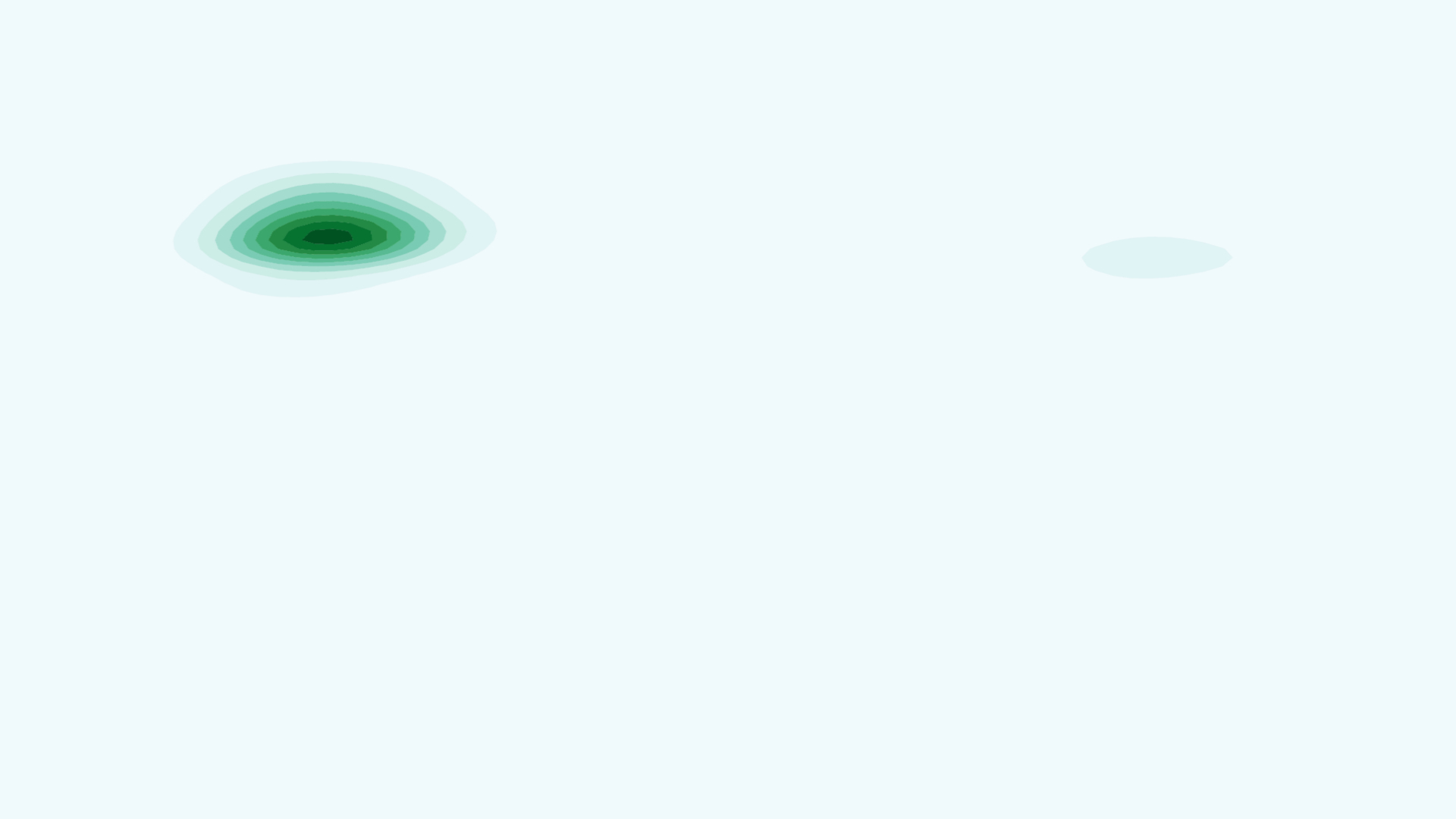}}
\subfigure[]{\includegraphics[width=0.3\linewidth]{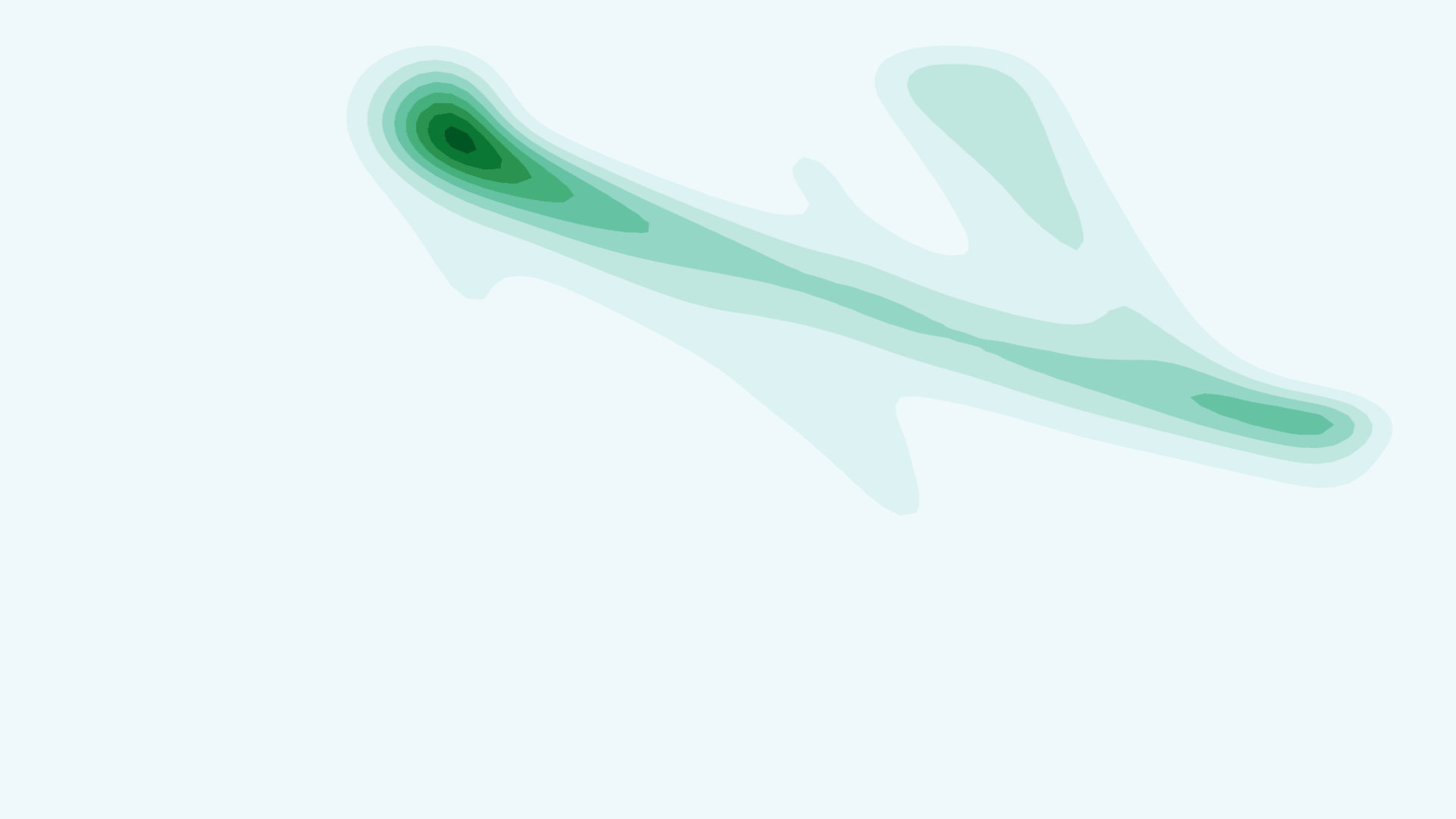}}
\subfigure[]{\includegraphics[width=0.3\linewidth]{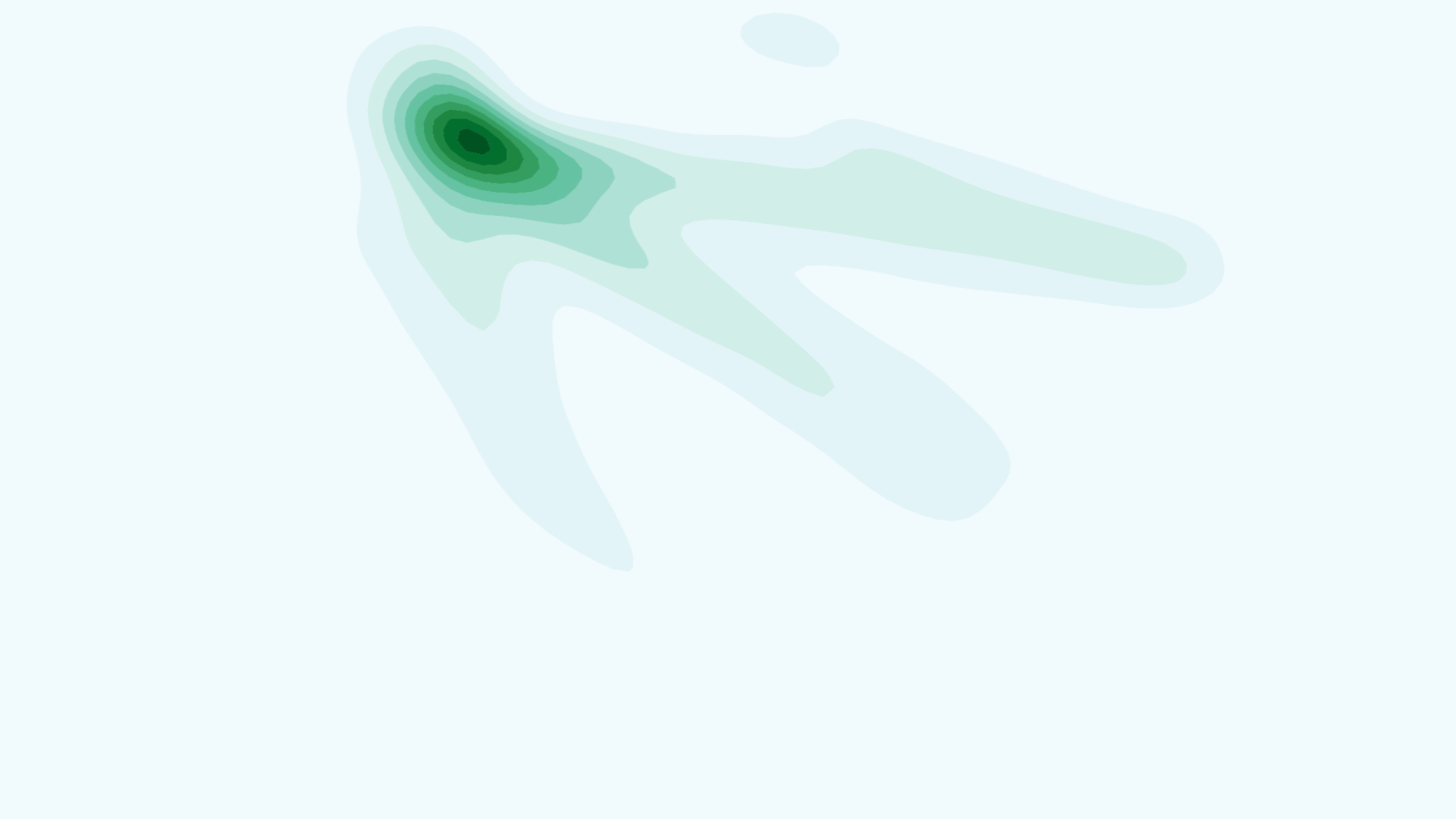}}

\caption{
Agent density maps of trajectory segments.
In each plot, we plotted the density maps of points of the segments corresponding to different agents using KDE.}
\label{fig:kde-density-clusters}
\end{figure}

Figure \ref{fig:kde-density-clusters}
shows the KDE density maps of segments corresponding to three agents.
In Figure \ref{fig:kde-density-clusters}(a)
segments gather the ticket counter at the left,
which means that the corresponding agent is assigned to short segments at the front
of the counter.
Figure \ref{fig:kde-density-clusters}(b)
shows many longer segments from the top-left entrance to the right exit, along with some shorter segments from the top-right entrance.
Therefore, this agent represents the dominant pedestrian flow from the top-left entrance.
In contrast,
Figure \ref{fig:kde-density-clusters}(c)
shows that the agent corresponds to pedestrian flows from three different directions
into the top-left exit (entrance).
This analysis is made possible using our proposed method for semantic
segmentation.
The original MDA provides clustering of trajectories only; therefore, the estimated agents are used
for analyzing entire trajectories, and not segments. RDP performs segmentation without semantics; thus 
segments cannot be classified.
On the other hand, the proposed method divides trajectories into segments and classifies segments using the
estimated agents, which enables us to perform
a behavior analysis by using trajectory segments.

\section{Conclusions}

In this paper, we proposed a semantic trajectory segmentation method
in which MDA and HMM are combined to estimate agent models and segment trajectories
according to the learned agents.
Experimental results using a dataset of real trajectories 
showed that 
the proposed method performs comparably with RDP, with only a small
difference in performance. Using  our improved MDA in the
proposed method greatly improves the performance compared to that of the
original MDA. Additionally, examples of the type of behavior analysis
that is made possible using the semantic segmentation results were also
provided.

\section*{Acknowledgments}

This work was supported by JSPS KAKENHI grant number JP16H06540.

\bibliographystyle{tADR}
\bibliography{bibtex_library}

\appendix

\section{$A$ and $b$}

By using the following formulas \cite{MatrixCookbook}
\begin{align}
\frac{\partial a^T X b}{X} & = a b^T\\
\frac{\partial a^T X^T b}{X} & = b a^T\\
\frac{\partial b^T X^T D X c}{X} & = D^T X b c^T + D X c b^T,
\end{align}
we have (note that we omit subscript $m$ from $A_m$ and $b_m$ for simplicity),
\begin{align}
\frac{\partial Q(\Theta, \hat\Theta)}{\partial A}
&=
\frac{\partial}{\partial A}
\sum_{k, h, t} \gamma^k
(x_t - A x_{t-1} - b)^T Q^{-1} (x_t - A x_{t-1} - b)
\\
&=
\frac{\partial}{\partial A}
\sum_{k, h, t}  \gamma^k (
x_t^T Q^{-1} x_t
- 2 b^T Q^{-1} x_t
+ b^T Q^{-1} b
\notag \\ & \phantom{=}
- 2 x_t^T Q^{-1} A x_{t-1}
+ 2 b^T Q^{-1} A x_{t-1}
+ x_{t-1}^T A^T Q^{-1} A x_{t-1})
\\
&=
\sum_{k, h, t}  \gamma^k (
- 2 Q^{-1} x_t x_{t-1}^T
+ 2 Q^{-1} b x_{t-1}^T
+ 2 Q^{-1} A x_{t-1} x_{t-1}^T )
= 0
\\
\sum_{k, h, t}  \gamma^k
x_t x_{t-1}^T
&=
\sum_{k, h, t}  \gamma^k (
A x_{t-1} x_{t-1}^T
+ b x_{t-1}^T)
\end{align}
and
\begin{align}
\frac{\partial Q(\Theta, \hat\Theta)}{\partial b}
&=
\frac{\partial}{\partial b}
\sum_{k, h, t}  \gamma^k
(x_t - A x_{t-1} - b)^T Q^{-1} (x_t - A x_{t-1} - b)
\\
&=
\sum_{k, h, t}  \gamma^k (
- 2 Q^{-1} x_t + 2 Q^{-1} b + 2 Q^{-1} A x_{t-1} )
= 0
\\
\sum_{k, h, t} \gamma^k
x_t
&=
\sum_{k, h, t} \gamma^k (
A x_{t-1} + b).
\end{align}

Therefore, we have the following system of equations.
\begin{align}
\sum_{k, h, t} \gamma^k x_t x_{t-1}^T &=  \sum_{k, h, t} \gamma^k (A x_{t-1} x_{t-1}^T + b x_{t-1}^T)
\\
\sum_{k, h, t} \gamma^k x_t &= \sum_{k, h, t} \gamma^k (A x_{t-1} + b)
\end{align}
By introducing the vectorization operator, we can rewrite it as
\begin{align}
\sum_{k, h, t} \gamma^k \vec(x_t x_{t-1}^T)
&=
\sum_{k, h, t} \gamma^k ((I_2 \otimes x_{t-1} x_{t-1}^T) \vec(A^T) + (I_2 \otimes x_{t-1}) b)
\\
\sum_{k, h, t} \gamma^k x_t &= \sum_{k, h, t} \gamma^k ((I_2 \otimes x_{t-1}^T) \vec(A^T) + b),
\end{align}
where $\vec$ is the vectorization operator and we used the following formula for $2 \times 2$ matrices $A$ and $B$.
\cite{MatrixAlgebra1997}
\begin{align}
\vec(AB) & = (I_2 \otimes A) \vec(B)
\\
\vec((AB)^T) & = \vec(B^T A^T) = (I_2 \otimes B^T) \vec(A^T).
\end{align}

Rewriting the system into matrix form, we have
\begin{align}
\sum_{k,h,t}\gamma^k
\begin{pmatrix}
(I_2 \otimes {\hat{\x}_{t-1}^k} (\hat{\x}_{t-1}^k){}^T) &
(I_2 \otimes \hat{\x}_{t-1}^k) \\
(I_2 \otimes (\hat{\x}_{t-1}^k){}^T) & I_2
\end{pmatrix}
\begin{pmatrix}
\vec(A^T) \\
\b
\end{pmatrix}
&=
\sum_{k,h,t}\gamma^k
\begin{pmatrix}
\vec(\hat{\x}_t^k (\hat{\x}_{t-1}^k){}^T)\\
\hat\x_{t}
\end{pmatrix}.
\end{align}

\section{$t_s$ and $t_e$}

By taking the derivative with respect to $\lambda_{sm}$, we have
\begin{align}
\frac{\partial Q(\Theta, \hat\Theta)}{\partial \lambda_{sm}}
&=
\frac{\partial}{\partial \lambda_{sm}}
\sum_{k,h/z=m} \gamma^k
\log \frac{\lambda_{sm}^{t_s^k}}{t_s^k !} e^{-\lambda_{sm}}
\\
&=
\frac{\partial}{\partial \lambda_{sm}}
\sum_{k,h/z=m} \gamma^k
(t_s^k \log \lambda_{sm}
- \log t_s^k !
-\lambda_{sm})
\\
&=
\sum_{k,h/z=m} \gamma^k
\left(\frac{t_s^k}{\lambda_{sm}} - 1 \right)
= 0
\\
\sum_{k,h/z=m} \gamma^k
&=
\sum_{k,h/z=m} \gamma^k
\frac{t_s^k}{\lambda_{sm}}
\\
\lambda_{sm}
&=
\frac{\sum_{k,h/z=m} \gamma^k t_s^k}{\sum_{k,h/z=m} \gamma^k}.
\end{align}

Similar equations are obtained for $\lambda_{em}$.

\end{document}